\documentclass[12pt,a4paper]{article} % 必须放在第一行
\usepackage[T1]{fontenc}
\usepackage{times}
\usepackage{astraxs} % 你的模板包

\usepackage{amssymb} 
\usepackage{amsmath}

\usepackage{algorithm}
\usepackage{algpseudocode}

\usepackage{cite}
\usepackage{graphicx}
\usepackage{caption}
\usepackage{multicol}
\usepackage{indentfirst}
\usepackage{geometry}
\usepackage{url}
\usepackage{adjustbox}
\usepackage{booktabs}

\begin{document}

\title{\fontsize{16}{18}\selectfont UAV-DETR: DETR for Anti-Drone Target Detection}
\authors{
Jun Yang$^1$$^{\rm *}$,Dong Wang$^1$, Hongxu Yin$^1$, Hongpeng Li$^1$, Jianxiong Yu$^1$ }
\footnotetext {\hskip -0.5cm\scriptsize $^{\rm *}$ Corresponding
author. E-mail: junyang@nwpu.edu.cn }

\affiliations{
$^1$School of Automation, NPU, Xian, ShaanXi  
%\\
%
%Seoul, South Korea
%%Department Aerospace Engineering, IIST, Thiruvananthapuram, Kerala 
\\
%$^2$Department, Institute address 2, India %
%\\
%$^3$Department of Engineering Mechanics, Seoul University, South Korea%
}

\maketitle
\begin{Abstract}
\textit{Drone detection is pivotal in numerous security and counter-UAV applications. However, existing deep learning-based methods typically struggle to balance robust feature representation with computational efficiency. This challenge is particularly acute when detecting miniature drones against complex backgrounds under severe environmental interference. To address these issues, we introduce UAV-DETR, a novel framework that integrates a small-target-friendly architecture with real-time detection capabilities. Specifically, UAV-DETR features a WTConv-enhanced backbone and a Sliding Window Self-Attention (SWSA-IFI) encoder, capturing the high-frequency structural details of tiny targets while drastically reducing parameter overhead. Furthermore, we propose an Efficient Cross-Scale Feature Recalibration and Fusion Network (ECFRFN) to suppress background noise and aggregate multi-scale semantics. To further enhance accuracy, UAV-DETR incorporates a hybrid Inner-CIoU and NWD loss strategy, mitigating the extreme sensitivity of standard IoU metrics to minor positional deviations in small objects. Extensive experiments demonstrate that UAV-DETR significantly outperforms the baseline RT-DETR on our custom UAV dataset (+6.61\% in $mAP_{50:95}$, with a 39.8\% reduction in parameters) and the public DUT-ANTI-UAV benchmark (+1.4\% in Precision, +1.0\% in F1-Score). These results establish UAV-DETR as a superior trade-off between efficiency and precision in counter-UAV object detection. The code is available at \url{https://github.com/wd-sir/UAVDETR}.
}

\end{Abstract}

%---------------------------1.引言-----------------------------------------------
\section{1. Introduction}
\setlength{\parindent}{10pt}
%This section contains the introduction, mainly covering the research significance and purpose, as well as the structure of the paper.
The rapid advancement and widespread deployment of unmanned aerial vehicles have brought significant convenience to civilian and commercial domains. However, the misuse of drones poses severe threats to public security, privacy, and critical infrastructure \cite{CHAMOLA_ATTACK,CNREVIEW,LOWSMALLUAV,REVIEW}. Consequently, the development of robust counter-UAV systems has become a critical security priority. Compared to traditional radar \cite{RADAR_MIMO, RADAR_TRACK} or radio frequency sensors \cite{RF_SIGNATURE, RFCNN}, vision-based drone detection offers competitive accuracy and reliability at substantially lower deployment costs \cite{DUT, VISION_BENCHMARK}, serving as the foundational step for subsequent tracking, interception, and combat intent recognition.

Despite the success of deep learning in general object detection, applying these algorithms directly to counter-UAV scenarios remains highly challenging due to the inherent visual characteristics of aerial targets. Drones often appear as extremely small targets in the visual field and exhibit drastic scale variations depending on their distance from the camera \cite{TIBNET}. Furthermore, real-world drone detection frequently suffers from severe background interference such as heavy cloud cover, mountainous terrain, and dense tree occlusion, making miniature targets easily confusable with environmental noise \cite{COMPLEX, DRONEVSBIRD}. Existing detection models often struggle to effectively balance high-resolution feature extraction with computational efficiency, a trade-off that is particularly critical for resource-constrained edge deployment \cite{PRUNED}. Standard convolution operations may lose critical high-frequency structural details of tiny targets during downsampling, while heavy transformer-based architectures incur prohibitive computational overhead for real-time applications. Additionally, standard evaluation metrics like intersection over union are highly sensitive to minor positional deviations of miniature bounding boxes, which severely degrades training stability\cite{INNERIOU}.

To address the aforementioned challenges, we propose \textbf{UAV-DETR}, a highly efficient and accurate real-time object detection framework tailored for counter-UAV operations. Built upon the real-time detection transformer architecture, our method systematically optimizes the structural design across the backbone, neck, and detection head. Specifically, as the backbone, we integrate Wavelet Transform Convolution (WTConv) into the basic blocks to formulate the WTConv Block. This precise integration preserves essential high-frequency spatial details and prevents information loss during downsampling. The neck architecture is constructed by cascading a Sliding Window Self-Attention Intra-scale Feature Interaction (SWSA-IFI) encoder and an Efficient Cross-Scale Feature Recalibration and Fusion Network (ECFRFN). These two components work synergistically within the neck to suppress background noise and aggregate multi-scale semantic features with minimal computational cost. Finally, the detection head employs a specialized hybrid loss function combining InnerCIoU and Normalized Wasserstein Distance (NWD) to significantly improve bounding box regression for tiny objects.

The main contributions of this paper are summarized as follows:
\begin{itemize}
	\item We propose UAV-DETR, a novel lightweight detection framework that significantly improves the detection accuracy of miniature drones in complex backgrounds while drastically reducing model parameters.
	\item We design a highly efficient neck architecture comprising the SWSA-IFI encoder and the ECFRFN. This cascaded neck design efficiently captures global context and recalibrates cross-scale features without incurring an excessive computational burden.
	\item We construct the WTConv Block by seamlessly incorporating WTConv into the backbone's basic blocks to enhance the retention of high-frequency structural details, and optimize the detection head by utilizing an InnerCIoU-NWD hybrid loss to alleviate the positional sensitivity of tiny targets.
	\item Extensive experiments on a custom UAV dataset and the public DUT-ANTI-UAV benchmark demonstrate that UAV-DETR achieves state-of-the-art performance, effectively breaking the bottleneck between high precision and lightweight deployment.
\end{itemize}

The remainder of this paper is organized as follows. Section 2 reviews related work on mainstream object detection methods and specific advancements in drone detection. Section 3 describes the methodology of the proposed UAV-DETR framework, including the overall architecture, customized feature extraction modules, and the improved loss function. Section 4 presents the experimental setup, evaluation metrics, and a comprehensive result analysis, covering generalization verification and ablation studies. In addition to quantitative metrics, this section features feature map and result visualizations for qualitative assessment, along with a critical discussion of algorithm failure cases. Finally, Section 5 concludes the paper.
%---------------------------2.相关研究-----------------------------------------------
\section{2. Related Work}
\setlength{\parindent}{10pt}

The landscape of object detection has been fundamentally reshaped by deep learning, shifting from hand-crafted feature extraction to data-driven feature learning through Convolutional Neural Networks. Early milestones were established by Ren et al., who introduced Faster R-CNN featuring a Region Proposal Network to generate high-quality object bounds \cite{FASTERRCNN}. Subsequently, Liu et al. proposed the Single Shot MultiBox Detector, which pioneered the use of multi-scale feature maps to achieve faster inference speeds \cite{SSD}. Building upon these one-stage foundations, the YOLO series has consistently dominated real-time object detection. Recent studies have explored YOLOv8 with an anchor-free design and decoupled heads \cite{YOLOV8}, while Wang et al. proposed YOLOv10 which successfully eliminated the non-maximum suppression step to reduce latency \cite{YOLOV10}. Successive iterations such as YOLO11 and YOLO12 have continued to optimize network topology and attention mechanisms \cite{YOLOV11, YOLOV12}, alongside highly customized variants like HyperYOLO designed for capturing complex high-order feature interrelationships \cite{HYPERYOLO}. To address the specific challenges of low-altitude aerial targets, researchers have also proposed specialized convolutional models such as PWM-YOLO and YOLO-GCOF, which incorporate customized feature extraction modules to improve drone detection accuracy \cite{PWMYOLO, YOLOGCOF}. Despite their remarkable efficiency across various applications, standard convolutional architectures inherently rely on progressive downsampling. This process often leads to the irreversible loss of high-frequency structural details, severely limiting their effectiveness when detecting extremely small aerial targets.

A significant paradigm shift occurred following the work of Vaswani et al. in 2017, who proposed the Transformer architecture relying entirely on self-attention mechanisms \cite{TRANSFORMER}. To adapt this architecture for computer vision, Dosovitskiy et al. introduced the Vision Transformer by transforming images into sequences of flattened patches \cite{VIT}. Building on this, Carion et al. proposed the Detection Transformer, framing object detection as a bipartite matching and direct set prediction problem \cite{DETR}. To address the slow convergence of the original model, Zhu et al. developed Deformable DETR by introducing deformable attention modules that focus only on sparse spatial locations \cite{DEFORMABLE_DETR}. Subsequent advancements led to highly optimized architectures such as DINO, which further refined query denoising and contrastive training for state-of-the-art performance \cite{DINO}. More recently, Zhao et al. introduced the Real-Time Detection Transformer to successfully bridge the gap between the high accuracy of attention mechanisms and strict real-time requirements \cite{RTDETR}. In aerial imagery, models like VRF-DETR demonstrate the strong applicability of self-attention mechanisms in extracting features from generic small objects, indicating great potential for the detection of miniature drones \cite{VRFDETR}. Similarly, novel methods like OSFormer pioneer the integration of small-object-friendly Transformers with a one-step detection paradigm to effectively suppress background noise and accentuate tiny targets \cite{OSFORMER}.

In real-world counter-UAV scenarios, ground-to-air vision-based systems face multifaceted challenges. Aerial targets typically exhibit extreme scale variations and often occupy merely a few pixels in distant captures \cite{REVIEW}. This extreme miniaturization makes them highly susceptible to severe background interference such as complex urban structures, dense foliage, or adverse illumination \cite{DRONEVSBIRD}. Furthermore, miniature drones are easily confused with background noise or other small airborne objects. To overcome these inherent difficulties and optimize small target detection, specific feature extraction mechanisms have been developed across diverse visual domains to preserve high-frequency details and handle scale variations. For instance, Finder et al. proposed wavelet convolutions to enlarge receptive fields efficiently \cite{WTCONV}. In image restoration tasks, omni-kernel networks have been introduced to learn comprehensive global-to-local feature representations \cite{OMINIKERNEL}. Spatial modeling has been further advanced through Shifted Window Self-Attention \cite{SWSA}. Additionally, context-guided spatial feature reconstruction explicitly extracts pyramid context for target modeling \cite{CGRFPN}, while Selective Boundary Aggregation shows remarkable promise in refining structural boundaries \cite{SBA}. Finally, advanced gradient paths like RepNCSPELAN4 are designed to optimize lightweight feature processing \cite{REPNCSPELAN4}. Inspired by these specialized developments, our proposed UAV-DETR is introduced to address the intricate challenges of aerial target perception. 
%---------------------------3.方法-----------------------------------------------
\section{3.Methodology}	
\setlength{\parindent}{0pt}

\subsection{3.1 Overall Architecture}
\setlength{\parindent}{10pt}

To address the unique challenges of drone detection from a counter-UAV perspective, we propose UAV-DETR, a novel detection framework built upon the robust foundation of the Real-Time Detection Transformer (RT-DETR) \cite{RTDETR}. As illustrated in Fig.~\ref{fig:overlook}, UAV-DETR is designed as an end-to-end pipeline that enhances multi-scale feature representation for miniature targets while maintaining real-time efficiency. 
\begin{center}
	\begin{minipage}{\textwidth}
		\centering
		\includegraphics[width=\linewidth]{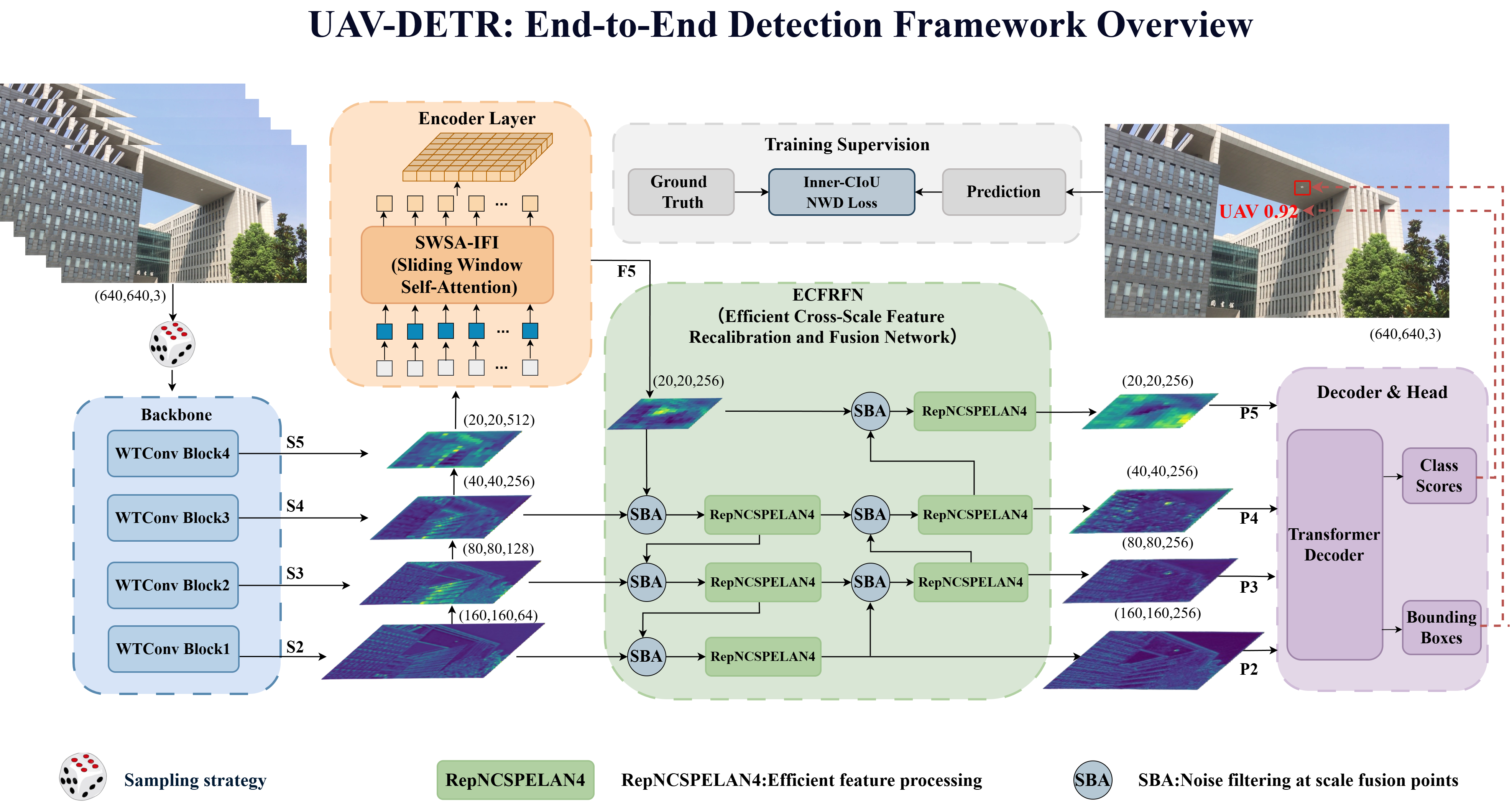}
		\captionof{figure}{The overall architecture of the proposed UAV-DETR. It consists of a WTConv-enhanced backbone, an SWSA-IFI encoder, an ECFRFN module, and a Transformer decoder supervised by InnerCIoU-NWD loss.}
		\label{fig:overlook}
	\end{minipage}
\end{center}

The overall architecture processes an input image through a continuous sequence of feature extraction, intra-scale encoding, cross-scale fusion, and decoding. Specifically, to mitigate the high visual redundancy and lack of variance among consecutive video frames, a random sampling strategy is employed during training, where one frame is randomly selected from every five. The sampled input image is then processed by a hierarchical backbone integrated with WTConv Blocks to extract multi-scale feature maps (denoted as $S_2$, $S_3$, $S_4$, and $S_5$). The highest-level semantic feature ($S_5$) is then passed through an intra-scale encoder featuring the SWSA-IFI module to efficiently capture global context. Subsequently, the resulting encoded feature ($F_5$), along with the shallower backbone features ($S_2$, $S_3$, $S_4$), are fed into the ECFRFN. Within this neck architecture, Selective Boundary Aggregation (SBA) and RepNCSPELAN4 modules work collaboratively to filter background noise and fuse features across different scales, yielding the refined multi-scale feature maps ($P_2$ to $P_5$). Finally, these aggregated features are processed by the Transformer Decoder to generate class scores and bounding box predictions, which are supervised during training by a customized InnerCIoU-NWD Loss. 

By unifying these components, UAV-DETR effectively balances the detection accuracy for visually fragmented drone targets with computational speed. The detailed formulations, structural mechanisms, and theoretical justifications for WTConv Block, SWSA-IFI, ECFRFN, and the InnerCIoU-NWD Loss are sequentially elaborated in Sections 3.2 through 3.5.

\subsection{3.2 WTConv Block}
\setlength{\parindent}{10pt}

Accurate detection of small UAVs in complex environments necessitates a feature extraction network capable of preserving fine-grained details while capturing global semantic dependencies. Standard Convolutional Neural Networks (CNNs), exemplified by ResNet-18, primarily rely on stacking small $3\times3$ kernels. While effective for general objects, this paradigm expands the Effective Receptive Field (ERF) slowly, which is suboptimal for small targets that occupy only a few pixels. Consequently, local operations tend to inadvertently amplify high-frequency background noise before meaningful semantic features are formed, leading to false positives or missed detections.

\begin{figure}[h]
	\centering
	\includegraphics[width=0.9\linewidth]{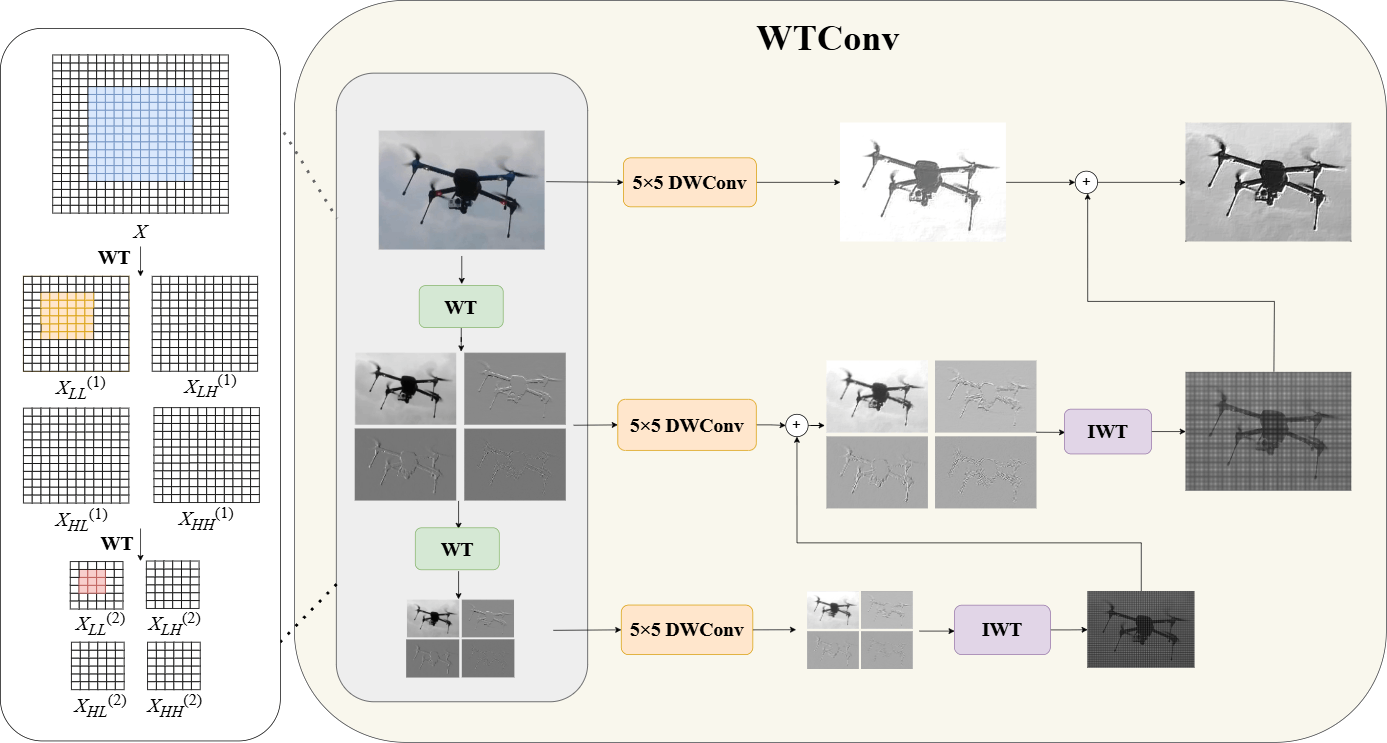}
	\caption{The process of the WTConv operation on a single channel, utilizing a 2-level wavelet decomposition and $5 \times 5$ kernel sizes for the depth-wise convolutions.}
	\label{fig:wtconv}
\end{figure}

To mitigate this, we introduce the WTConv Block to construct a frequency-aware backbone. Leveraging the Multi-Resolution Analysis (MRA) property of wavelets, WTConv enables the network to respond to low-frequency components—corresponding to object shapes—over a rapidly expanding spatial range. This effectively suppresses high-frequency noise while enhancing the structural representation of small UAVs. As illustrated in Fig.~\ref{fig:wtconv}, the WTConv module achieves this via a cascade of wavelet decomposition and frequency-domain interaction. Given an input feature map $X_{in}$, we employ the 2D Haar Wavelet Transform (WT) to recursively decompose it into four sub-bands: the low-frequency approximation $X_{LL}$ and high-frequency details $\{X_{LH}, X_{HL}, X_{HH}\}$ for $L$ levels. This cascading process generates a feature pyramid where deeper levels correspond to lower frequencies and exponentially larger receptive fields. To facilitate interaction across these scales, we concatenate the sub-bands at each level $i$ and apply a depth-wise convolution. This operation is formulated as:
\begin{equation}
	Y^{(i)} = \mathcal{S}^{(i)}\left( \text{DWConv}_{5 \times 5} \left(\text{Concat}[X_{LL}^{(i)}, X_{LH}^{(i)}, X_{HL}^{(i)}, X_{HH}^{(i)}] \right) \right),
\end{equation}
where $\text{DWConv}_{5 \times 5}$ denotes a depth-wise convolution with a $5 \times 5$ kernel to process frequency information efficiently, and $\mathcal{S}^{(i)}$ represents a learnable channel-wise scaling factor. Following the convolution, an Inverse Wavelet Transform (IWT) is employed for recursive reconstruction. Crucially, we introduce an additive fusion strategy where the global structural context from deeper levels flows back to guide the feature reconstruction at shallower levels. The reconstruction at level $i$ is defined as:
\begin{equation}
	\hat{X}_{LL}^{(i)} = \text{IWT} \left( Y_{LL}^{(i)} + \hat{X}_{LL}^{(i+1)}, Y_{LH}^{(i)} Y_{HL}^{(i)}, Y_{HH}^{(i)} \right),
\end{equation}
where $\hat{X}_{LL}^{(L+1)} = 0$. The term $Y_{LL}^{(i)} + \hat{X}_{LL}^{(i+1)}$ ensures that the low-frequency information is progressively enhanced by the global context captured at deeper levels.

\begin{figure}[h]
	\centering
	\includegraphics[width=\linewidth]{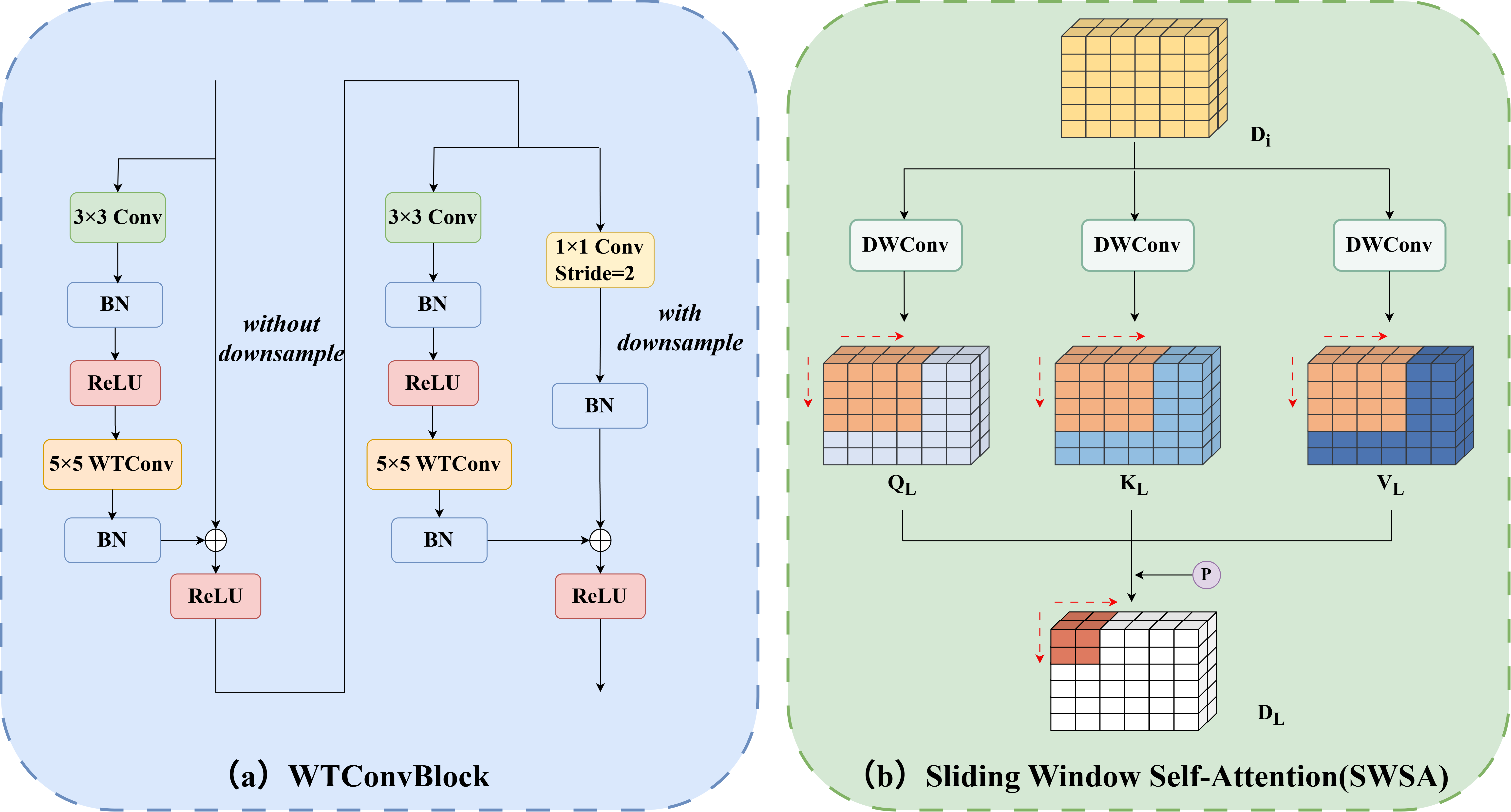}
	\caption{(a) The detailed architecture of the proposed WTConv Block. (b) The structural principle of the SWSA attention mechanism.}
	\label{fig:wtconv_swsa}
\end{figure}

Building upon this mechanism, we propose the WTConv Block as the fundamental building unit of our backbone, as illustrated in Fig.~\ref{fig:wtconv_swsa}(a). Distinct from standard residual blocks, we formulate the WTConv Block as a composite module consisting of two cascaded stages: a semantic refinement stage (without downsampling) followed by a spatial compression stage (with downsampling). In each stage, we modify the standard ResNet architecture by retaining the initial $3 \times 3$ convolution to capture local texture cues, while replacing the subsequent convolution with the frequency-aware WTConv module to expand the receptive field. Formally, let $x$ denote the input feature map. The feature propagation is defined as a two-step process. In the first stage, the feature is refined at the original resolution via $x' = \sigma\left( \mathcal{F}(x) + x \right)$, where $\sigma$ denotes the ReLU activation and $\mathcal{F}(\cdot)$ represents the residual mapping function:
\begin{equation}
	\mathcal{F}(x) = \text{BN}\left( \text{WTConv}\left( \sigma(\text{BN}(\text{Conv}_{3 \times 3}(x))) \right) \right),
\end{equation}

Subsequently, the refined intermediate feature $x'$ serves as the input for the downsampling stage to generate the final output $y$:
\begin{equation}
	y = \sigma\left( \mathcal{F}_{s=2}(x') + \text{BN}(\text{Conv}_{1 \times 1,s=2}(x')) \right),
\end{equation}
where $\mathcal{F}_{s=2}$ denotes the residual mapping with a stride of 2, and the term $\text{BN}(\text{Conv}_{1 \times 1,s=2}(x'))$ represents the projection shortcut for spatial downsampling. This cascaded design establishes a dual-pathway mechanism: the first stage prioritizes the preservation of local details and background noise filtering, while the second stage focuses on encoding global structural integrity into a compact representation.

%-------------------------特征融合网络--------------------------------------
\subsection{3.3 Feature Encoding and Fusion Neck}
\setlength{\parindent}{10pt}

Following the hierarchical feature extraction by the WTConv-enhanced backbone, it is essential to establish global context and effectively fuse multi-scale features to accommodate the extreme scale variations of UAVs. To achieve this, we design a comprehensive intermediate processing architecture composed of two sequential components: an intra-scale feature encoder and a cross-scale fusion network. First, the deepest semantic features output by the backbone are processed to capture global dependencies. Subsequently, these enriched high-level features are aggregated with shallower, high-resolution feature maps to construct a robust multi-scale representation, effectively filtering out background noise in the process. The specific designs of these two core components, namely SWSA-IFI and ECFRFN, are detailed in the following subsections.

%-------------------------编码器--------------------------------------
\subsubsection{3.3.1 SWSA-IFI Encoder}
\setlength{\parindent}{10pt}

While the standard RT-DETR leverages the Attention-based Intra-scale Feature Interaction (AIFI) module to capture global semantic dependencies on high-level feature maps, it faces limitations in small UAV detection. Small targets typically occupy very few pixels, and standard global self-attention—which computes dependencies across the entire image—often introduces excessive background noise. This global context can overshadow the weak feature representations of small targets. To mitigate this and enhance the model's focus on local contextual information, we propose replacing the standard encoder layer in AIFI with a Sliding Window Self-Attention (SWSA) mechanism.

SWSA is designed to restrict attention computation to a localized region, thereby reducing computational redundancy while preserving fine-grained details. Architecturally, the SWSA decomposes the transformer block into a Token Mixer and a Channel Mixer, as shown in Fig.~\ref{fig:wtconv_swsa}(b). Unlike standard multi-head attention that relies on dense linear layers, the Token Mixer facilitates local feature projection by employing $1\times 1$ depth-wise convolutions to generate query ($Q$), key ($K$), and value ($V$) matrices. Operating essentially as an independent per-channel scalar multiplication, this specific convolutional design drastically reduces parameter redundancy compared to standard projections. Furthermore, the attention mechanism operates within a sliding window defined by size $w$ and stride $s$. By ensuring $w > s$, the overlapping windows promote information flow across boundaries, thereby preserving spatial continuity.

Since self-attention mechanisms are inherently permutation-invariant, explicit spatial priors are requisite. We incorporate a learnable Relative Positional Encoding (RPE), denoted as $P_{rel}$. The attention output within a window is computed as:
\begin{align}
	Q, K, V &= \text{DWConv}_{1\times 1}(X), \label{eq:swsa_qkv} \\
	\text{Attention}(Q, K, V) &= \text{Softmax}\left(\frac{QK^\top}{\sqrt{d_k}} + P{rel}\right)V, \label{eq:swsa_attn}
\end{align}
where $X$ represents the input feature map and $d_k$ is the scaling factor. $P_{rel}$ enables the network to learn the spatial arrangement of pixels, which is critical for distinguishing the structural details of small UAVs. Following local aggregation and residual addition, a Channel Mixer facilitates cross-channel information exchange. We implement this via a Convolutional Feed-Forward Network (FFN) comprising two $1\times1$ convolutional layers. To ensure stable training and align with the optimized inference architecture, the final output is obtained after a subsequent residual connection and layer normalization (LN), formulated as follows:
\begin{equation}
	O' = \text{LN}(\text{FFN}(O) + O) = \text{LN}(\text{Conv}_{1\times 1}(\sigma(\text{Conv}_{1\times 1}(O))) + O), \label{eq:swsa_ffn}
\end{equation}
where $\sigma$ denotes the activation function (e.g., GELU) and $O$ represents the output from the preceding Token Mixer. In the proposed SWSA-IFI module, we replace the standard Transformer Encoder Layer with this SWSA-based architecture. High-level feature maps are first processed by the Token Mixer (incorporating the sliding window and RPE), followed by the convolutional Channel Mixer. This design enables the model to efficiently capture pixel-level relationships within spatially adjacent regions, effectively filtering background noise while highlighting miniature UAV targets.
	
%-------------------------高效融合--------------------------------------
\subsubsection{3.3.2 ECFRFN Module}
\setlength{\parindent}{10pt}

\begin{figure}[h]
	\centering
	\includegraphics[width=\linewidth]{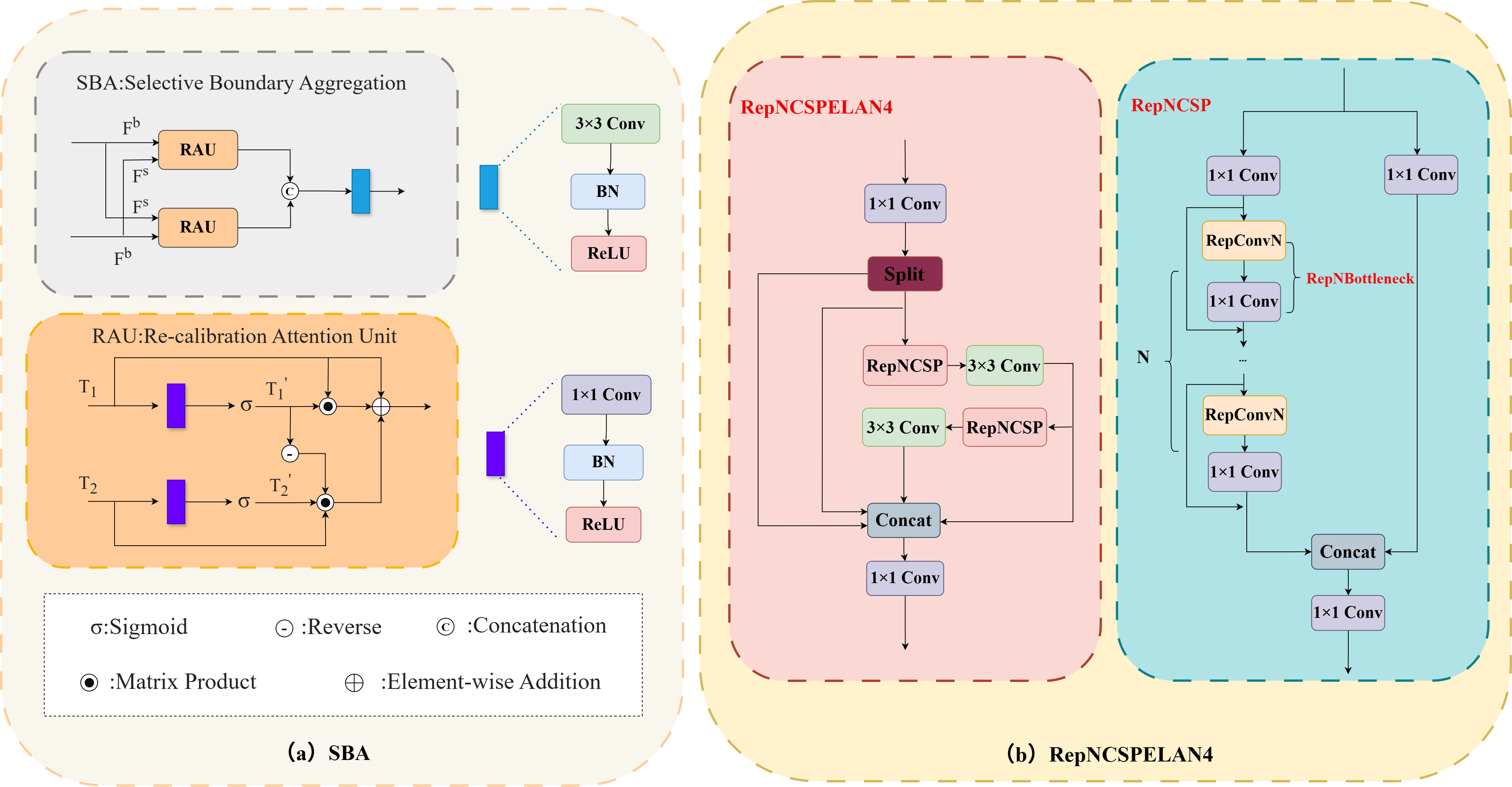}
	\caption{(a) The architecture of the Selective Boundary Aggregation (SBA) module. (b) The schematic of the RepNCSPELAN4 module illustrating the structural re-parameterization mechanism.}
	\label{fig:sba_rep}
\end{figure}

Addressing the scale variation inherent in small UAV detection requires a robust mechanism to integrate fragmented features. Naive concatenation of hierarchical features often results in semantic ambiguity and redundancy, as deep semantic maps and shallow detail maps possess distinct distributions. To resolve this, we propose the ECFRFN as the detector's fusion neck. As depicted in Fig.~\ref{fig:sba_rep}, the ECFRFN functions as an advanced feature pyramid, seamlessly aggregating high-level semantics with fine-grained spatial details. The architecture is distinguished by two strategic components: the SBA module, designed for precise feature alignment, and the RepNCSPELAN4 module, which ensures computational efficiency without compromising representational depth.

% 【SBA: 强调对齐问题和边界重要性】
\textbf{SBA Module.} 
Conventional feature fusion mechanisms, typically relying on linear upsampling followed by element-wise addition, frequently suffer from spatial misalignment caused by the semantic gap between scales. This misalignment is particularly detrimental for small UAVs, where boundary blurring can lead to severe detection failures against complex backgrounds. To mitigate this, we introduce the SBA module to adaptively recalibrate feature responses prior to fusion. As illustrated in Fig.~\ref{fig:sba_rep}(a), the SBA incorporates a Re-calibration Attention Unit (RAU) that explicitly models the dependencies between boundary delineation and internal texture. By dynamically weighting the input features, the RAU suppresses background noise amplification during upsampling while enhancing the structural integrity of the target. This ensures that only the most discriminative cues are propagated to the subsequent detection head.

% 【RepNCSPELAN4: 强调梯度流优势和重参数化的解耦特性】
\textbf{RepNCSPELAN4.} 
Balancing detection accuracy with real-time inference speed remains a critical bottleneck for UAV detectors deployed in resource-constrained edge environments. We address this limitation by replacing standard convolutional blocks with the proposed RepNCSPELAN4 module within the feature aggregation path. As illustrated in Fig.~\ref{fig:sba_rep}(b), this architecture synergizes the efficient gradient flow inherent to Cross Stage Partial networks with the robust layer aggregation capabilities of ELAN. Crucially, we introduce structural re-parameterization into the bottleneck layers. This paradigm effectively decouples the training and inference architectures: during optimization, the module leverages a multi-branch topology to capture diverse feature representations, whereas for deployment, these constituent branches are algebraically fused into a single $3\times3$ convolution. Consequently, this design drastically reduces both the total parameter count and floating-point operations while maintaining a rich gradient flow, rendering the ECFRFN highly optimized for latency-sensitive hardware execution.

%-------------------------损失函数--------------------------------------
\subsection{3.4 InnerCIoU-NWD Hybrid Loss}
\setlength{\parindent}{10pt}

In the context of UAV detection, targets are typically characterized by their minuscule scale and complex backgrounds. Traditional loss functions like GIoU rely heavily on the geometric overlap between the predicted box and the ground truth. However, for small objects, this approach exhibits distinct limitations: it is highly sensitive to positional deviations—where a shift of a few pixels causes a drastic drop in IoU—and suffers from slow convergence when the predicted box is enclosed within the ground truth.

To address these challenges, we propose a hybrid loss function that synergizes NWD and Inner-CIoU. Since small objects often lack sufficient appearance information, purely geometric overlap is insufficient. We adopt the NWD metric to model bounding boxes as 2D Gaussian distributions rather than rigid rectangles. For a bounding box $B = (cx, cy, w, h)$, modeled as $\mathcal{N}(\mu, \Sigma)$, the similarity between the prediction $A$ and ground truth $B$ is measured by the Wasserstein distance:
\begin{equation}
	W_2^2(\mathcal{N}_A, \mathcal{N}_B) = \|\mu_A - \mu_B\|_2^2 + \|\Sigma_A^{1/2} - \Sigma_B^{1/2}\|_F^2, \label{eq:wass_dist}
\end{equation}
Accordingly, the NWD loss is formulated as:
\begin{equation}
	\mathcal{L}_{NWD} = 1 - \exp\left(-\frac{\sqrt{W_2^2(\mathcal{N}_A \mathcal{N}_B)}}{C}\right), \label{eq:nwd_loss}
\end{equation}
where $C$ is a dataset-specific constant. The probabilistic nature of NWD ensures that even non-overlapping boxes yield non-zero gradients, providing a continuous learning signal crucial for tiny targets.

While NWD ensures robustness, high-precision localization requires further optimization. To this end, we substitute the standard CIoU with Inner-CIoU, which employs an auxiliary bounding box scaled by a factor $r$. For a given bounding box $B = (cx, cy, w, h)$, the auxiliary inner box is generated by scaling its width and height while preserving the center coordinates as $B_{inner} = (cx, cy, r \cdot w, r \cdot h)$. To construct the optimization objective, we build upon the standard CIoU metric. The foundational geometric overlap ($IoU$) and the comprehensive CIoU loss are logically formulated as:
\begin{align}
	IoU &= \frac{|B^p \cap B^{gt}|}{|B^p \cup B^{gt}|}, \label{eq:iou_def} \\
	\mathcal{L}_{CIoU}(B^p, B^{gt}) &= 1 - IoU + \frac{\rho^2(b^p, b^{gt})}{c^2} + \alpha v, \label{eq:ciou_base}
\end{align}
where $B^p$ and $B^{gt}$ represent the predicted and ground truth boxes, respectively. The term $\rho(\cdot)$ denotes the Euclidean distance between their central points $b^p$ and $b^{gt}$, and $c$ is the diagonal length of the smallest enclosing box. The parameter $v = \frac{4}{\pi^2}(\arctan\frac{w^{gt}}{h^{gt}} - \arctan\frac{w^p}{h^p})^2$ measures the aspect ratio consistency, and $\alpha = \frac{v}{(1 - IoU) + v}$ serves as a dynamic trade-off weight.

By strictly applying this comprehensive geometric constraint to the localized inner regions, Inner-CIoU amplifies the effective gradient in high-IoU scenarios, thereby accelerating convergence. The specific loss is formulated by substituting the scaled boxes into Eq. (\ref{eq:ciou_base}), defined as Eq. (\ref{eq:inner-ciou}). Finally, to balance robustness and precision, the total bounding box regression loss $\mathcal{L}_{box}$ is constructed by integrating both components as Eq. (\ref{eq:total_box_loss}):
\begin{align}
	\mathcal{L}_{Inner-CIoU} = \mathcal{L}_{CIoU}(B^{p}_{inner}, B^{gt}_{inner}),\label{eq:inner-ciou}\\
	\mathcal{L}_{box} = \lambda \cdot \mathcal{L}_{Inner-CIoU} + (1 - \lambda) \cdot \mathcal{L}_{NWD}, \label{eq:total_box_loss}
\end{align}
where $\lambda$ is a hyperparameter regulating their relative contributions. This combined strategy significantly enhances detection performance for small-scale UAVs compared to the baseline.
%-------------------------伪代码--------------------------------------

\subsection{3.5 Pseudo Code}
\setlength{\parindent}{10pt}
%The pseudo code presented in Algorithm~\ref{alg:proposed_method} provides an algorithmic depiction of the proposed UAV-DETR. This method encompasses several key processes:
The algorithmic implementation of the proposed UAV-DETR is systematically detailed in Algorithm~\ref{alg:proposed_method}. This end-to-end optimization pipeline encompasses four key processes:
\begin{algorithm}[h]
	\caption{Training Scheme of UAV-DETR}
	\label{alg:proposed_method}
	\begin{algorithmic}[1]
		\Require Input image $\mathbf{I}$, Ground Truth $\mathcal{B}_{gt}, \mathcal{C}_{gt}$
		\Ensure Optimized Network Parameters $\Theta$
		
		\State \textbf{Step 1: Frequency-aware Feature Extraction}
		\State Initialize backbone features $\mathcal{F} = \emptyset$, $x \leftarrow \mathbf{I}$
		\For{$i \in \{2, 3, 4, 5\}$}
		\State $x \leftarrow \text{WTConv Block}_i(x)$ \Comment{Extract multi-scale features}
		\State $\mathcal{F} \leftarrow \mathcal{F} \cup \{F_i\}$ \Comment{Save feature map $F_i$ at stride $2^i$}
		\EndFor
		
		\State \textbf{Step 2: Global Context Enhancement}
		\State $F_5' \leftarrow \text{SWSA-IFI}(F_5)$ \Comment{Sliding Window Self-Attention}
		
		\State \textbf{Step 3: Cross-Scale Recalibration and Fusion}
		\State $P_5 \leftarrow \text{RepNCSPELAN4}(\text{SBA}(F_5'))$ 
		\For{$i \in \{4, 3, 2\}$} \Comment{Top-down pathway}
		\State $F_{aligned} \leftarrow \text{SBA}(\text{Upsample}(P_{i+1}), F_i)$ \Comment{Selective Boundary Aggregation}
		\State $P_i \leftarrow \text{RepNCSPELAN4}(F_{aligned})$ \Comment{Efficient feature processing}
		\EndFor
		
		\State \textbf{Step 4: Prediction and Hybrid Optimization}
		\State $H_{feat} \leftarrow \text{TransformerDecoder}(\{P_2, \dots, P_5\})$
		\State $\mathcal{B}_{pred}, \mathcal{C}_{pred} \leftarrow \text{DetectionHeads}(H_{feat})$
		\State $\mathcal{L}_{reg} \leftarrow \alpha \cdot \text{InnerCIoU} + (1-\alpha) \cdot \text{NWD}$ \Comment{Hybrid geometric \& distribution loss}
		\State Update $\Theta$ via backpropagation of $\mathcal{L}_{cls} + \mathcal{L}_{reg}$
		
		\State \Return $\Theta$
	\end{algorithmic}
\end{algorithm}

\textbf{Frequency-aware Feature Extraction:} The process begins with extracting multi-scale representations using a backbone constructed with WTConv Block. This involves decomposing input features into frequency sub-bands to separately process structural shapes and high-frequency details. The frequency-aware mechanism is essential for preserving the fine-grained integrity of small UAVs while effectively suppressing background noise interference.

\textbf{Global Context Enhancement:} The deepest feature maps are subsequently processed by the SWSA-IFI encoder to capture long-range dependencies. This involves partitioning features into windows to aggregate global semantic context. The proposed attention mechanism is crucial for enriching the feature representation of small targets, facilitating their distinction from complex, cluttered environments.

\textbf{Cross-Scale Recalibration and Fusion:} To alleviate the semantic gap between different scales, the ECFRFN is applied to the extracted hierarchical features. This process incorporates the SBA for precise feature alignment and the RepNCSPELAN4 for lightweight processing. This configuration allows for effective feature calibration, optimizing computational efficiency without sacrificing detection accuracy.

\textbf{Prediction and Hybrid Optimization:} With the calibrated feature pyramid, a Transformer Decoder is utilized to generate final object predictions. This phase introduces a hybrid loss function combining Inner-CIoU and NWD for joint optimization. The hybrid strategy is critical for ensuring sensitivity to tiny objects and delivering precise geometric localization during the training process.
%-------------------------实验--------------------------------------
\section{4.Experiments}
\vspace*{-10pt}
\setlength{\parindent}{10pt}
%-------------------------数据集--------------------------------------
\subsection{4.1 Dataset Preparation}
\setlength{\parindent}{10pt}

In recent years, several anti-UAV datasets have been introduced to advance vision-based detection, including  the DUT Anti-UAV dataset \cite{DUT}, the TIB dataset \cite{TIBNET}, the UAVSwarm dataset \cite{UAVSWARM}, and DroneMMset. However, these publicly available datasets often emphasize specific and isolated challenges. For instance, the UAVSwarm dataset primarily focuses on multi-target tracking within drone swarms, while the TIB dataset is specifically tailored for extremely small aerial targets. Similarly, the DUT dataset emphasizes multi-scenario variations, and DroneMMset predominantly addresses severe illumination changes. While valuable, relying solely on these specialized datasets may not fully reflect the compounded complexities of real-world counter-UAV operations. 

\begin{figure}[h]
	\centering
	\includegraphics[width=\linewidth]{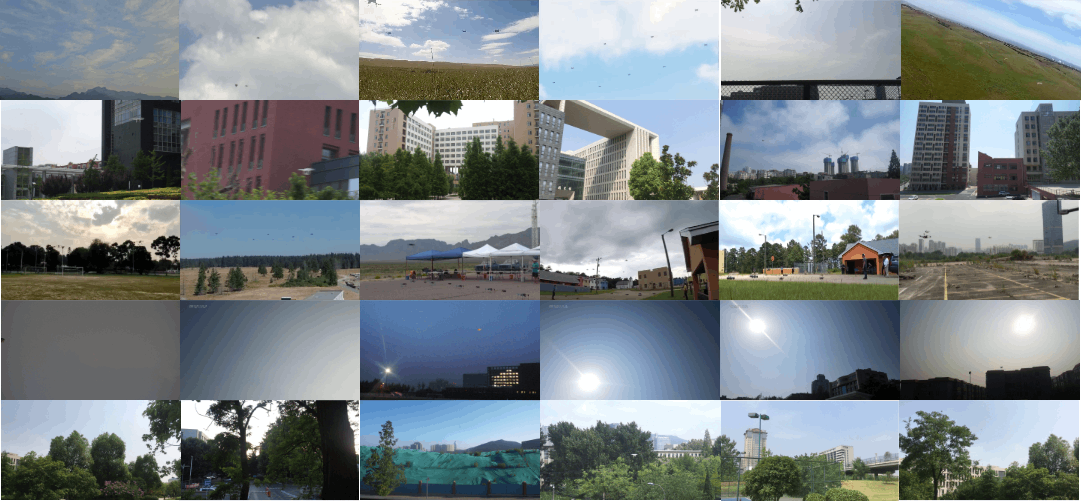}
	\caption{Sample images of UAV dataset.}
	\label{fig:uavdataset}
\end{figure}

Therefore, to rigorously evaluate the robustness and versatility of the proposed method against compounded environmental challenges, we constructed a comprehensive UAV detection dataset. As visualized in Fig.~\ref{fig:uavdataset}, this dataset encompasses a wide spectrum of environmental variability, incorporating diverse background clutter such as urban skylines and foliage alongside varying illumination, weather conditions, and scenarios featuring both single and multiple UAVs at drastic scale variations. The data sources are a hybrid integration of existing open-source archives and self-collected real-world footage. Crucially, to address the issue of high temporal redundancy inherent in video-based data where adjacent frames exhibit excessive visual similarity that consumes training resources without adding information gain, we implemented a temporal subsampling strategy. Specifically, for every sequence of five adjacent frames, a single representative image is randomly extracted to preserve feature diversity while reducing computational overhead. After rigorous cleaning and subsampling, the final curated dataset comprises a total of 14,713 images. Finally, the curated dataset is randomly partitioned into training, validation, and testing subsets following a 7:2:1 ratio to ensure a fair and comprehensive performance assessment.

%-------------------------实验设置--------------------------------------
\subsection{4.2. Implementation Details and Experimental Setup}
\setlength{\parindent}{10pt}

To ensure a fair and consistent evaluation, all experiments were conducted on a uniform hardware and software platform. The deep learning models were implemented using the PyTorch framework and executed on a high-performance server equipped with an NVIDIA RTX 3090 GPU. The specific hardware and software configurations are detailed in Table \ref{tab:setup}.

In terms of model selection, we employed 11 representative state-of-the-art baselines to benchmark against the proposed UAV-DETR. Fundamentally, our network operates as a universal data-driven detection framework that learns robust target features autonomously, ensuring that direct performance comparisons with general detectors are highly relevant and methodologically sound. Furthermore, to guarantee a strictly fair quantitative evaluation, these baseline models were carefully selected based on their comparable parameter counts and computational complexity (FLOPs). The comprehensive selection includes:
\begin{itemize}
	\item \textbf{CNN-based architectures:} The classic two-stage Faster R-CNN and single-stage SSD, both equipped with standard ResNet-50 backbones, alongside the state-of-the-art YOLO series (YOLOv8m, YOLOv10m, YOLO11m, and YOLO12m) and an improved YOLO variant, Hyper-YOLOm.
	\item \textbf{Transformer-based architectures:} The standard DETR and Deformable DETR, both utilizing ResNet-50 backbones, the baseline RT-DETR configured with a lightweight ResNet-18 backbone, and the recent VRF-DETR.
\end{itemize}

All models were trained for 100 epochs to balance convergence speed and computational resource utilization. To rigorously evaluate the feature extraction capability of the architectures themselves—rather than the benefits of transfer learning from large-scale datasets like COCO—the primary experimental protocol involves training all models from scratch (i.e., without loading pre-trained weights). However, empirical observations indicate that certain earlier architectures, specifically Faster R-CNN, SSD, DETR, and Deformable DETR, exhibit significant convergence difficulties and suboptimal performance when trained from scratch on this specific dataset. To address this and establish competitive baselines, we conducted separate experiments for these four models initialized with pre-trained weights, denoted with the subscript PT (e.g., Faster R-CNN$_{\text{PT}}$, DETR$_{\text{PT}}$).

\begin{table}[ht]
	\centering
	\caption{Implementation Environment Details}
	\label{tab:setup}
	\renewcommand{\arraystretch}{1.2} % 保持你设置的行高，增加呼吸感
	\setlength{\tabcolsep}{15pt} % 【核心魔法】：把两列的间距稍微撑开一点，显得大方而不空旷
	\begin{tabular}{lc} 
		\toprule[1.2pt]
		\textbf{Component} & \textbf{Specification} \\
		\midrule[0.5pt]
		Operating System & Ubuntu 20.04.6 LTS \\
		CPU & 12th Gen Intel(R) Core(TM) i7-12700KF @ 3.60 GHz \\
		RAM & 64 GB \\
		GPU & NVIDIA RTX3090 (24 GB) \\
		\midrule
		Python Version & 3.9.25 \\
		Deep Learning Framework & PyTorch 1.12.1 \\
		CUDA Version & 11.3 \\
		cuDNN Version & 8.3.2 \\
		\bottomrule[1.2pt]
	\end{tabular}
\end{table}

\subsection{4.3 Evaluation Metrics}
\setlength{\parindent}{10pt}

To conduct a comprehensive quantitative analysis of the proposed UAV-DETR, we employ a multi-dimensional evaluation protocol covering both detection accuracy and computational efficiency. To clearly denote the optimization direction of each metric, we use ($\uparrow$) to indicate that higher values are preferred, and ($\downarrow$) to indicate that lower values are better.

\textbf{Detection Performance Metrics.} Following standard benchmarks such as COCO and PASCAL VOC, we utilize Precision ($P$, $\uparrow$), Recall ($R$, $\uparrow$), and F1-score ($F1$, $\uparrow$) to evaluate the basic classification and localization capabilities. These are defined as:
\begin{equation}
	P = \frac{TP}{TP + FP}, \quad R = \frac{TP}{TP + FN}, \quad F1 = \frac{2 \times P \times R}{P + R},
\end{equation}
where $TP$, $FP$, and $FN$ denote True Positives, False Positives, and False Negatives, respectively. To further assess the robustness of the detector under varying overlap thresholds, we adopt the Mean Average Precision (mAP, $\uparrow$), which represents the area under the Precision-Recall curve averaged across all classes. We report three specific mAP variants:
\begin{itemize}
	\item \textbf{mAP$_{\mathbf{50}}$ ($\uparrow$):} The mAP calculated at a single Intersection over Union (IoU) threshold of 0.5. This metric primarily reflects the model's ability to roughly locate objects.
	\item \textbf{mAP$_{\mathbf{75}}$ ($\uparrow$):} The mAP at a stricter IoU threshold of 0.75, which demands higher localization precision.
	\item \textbf{mAP$_{\mathbf{50:95}}$ ($\uparrow$):} The average mAP over 10 IoU thresholds ranging from 0.5 to 0.95 with a step size of 0.05. This is the most rigorous metric, particularly critical for small UAV detection, where high-IoU matching is exceedingly challenging due to the minuscule pixel area of the targets.
\end{itemize}

\textbf{Computational Efficiency Metrics.} Beyond detection accuracy, evaluating the model's lightweight characteristics is essential given the constraints of deploying UAV detectors on resource-limited hardware, typified by edge computing devices. To this end, we incorporate two key efficiency indicators:
\begin{itemize}
	\item \textbf{Parameters (Params, $\downarrow$):} Measured in millions (M), this metric quantifies the spatial complexity and storage requirements of the model.
	\item \textbf{Floating Point Operations (FLOPs, $\downarrow$):} Measured in giga-floating point operations (G), this metric evaluates the time complexity and computational cost during inference.
\end{itemize}

Ultimately, minimal values in Params and FLOPs ($\downarrow$), coupled with maximized mAP scores ($\uparrow$), demonstrate the proposed method's superiority in achieving an optimal trade-off between detection accuracy and deployment efficiency.

%-------------------------实验结果（对比实验）--------------------------------------
\subsection{4.4 Experimental Results}
\setlength{\parindent}{10pt}
%我们在数据集uavdataset上对上述12个模型进行训练，统一在测试集进行评估，得到结果。F1-C和PR曲线如图1所示.可以看出：（描述实验结果），还有实验指标如表1所示，（定量描述）。
To empirically validate the effectiveness of the proposed method, we conducted a comparative analysis against 11 state-of-the-art detectors on the constructed UAV dataset. All models were evaluated on the test set following the identical training protocol described in Section 4.2.

\subsubsection{4.4.1. Qualitative Analysis}
\begin{figure}[ht]
	\centering
	\begin{minipage}{0.49\linewidth}
		\centering
		\includegraphics[width=\linewidth]{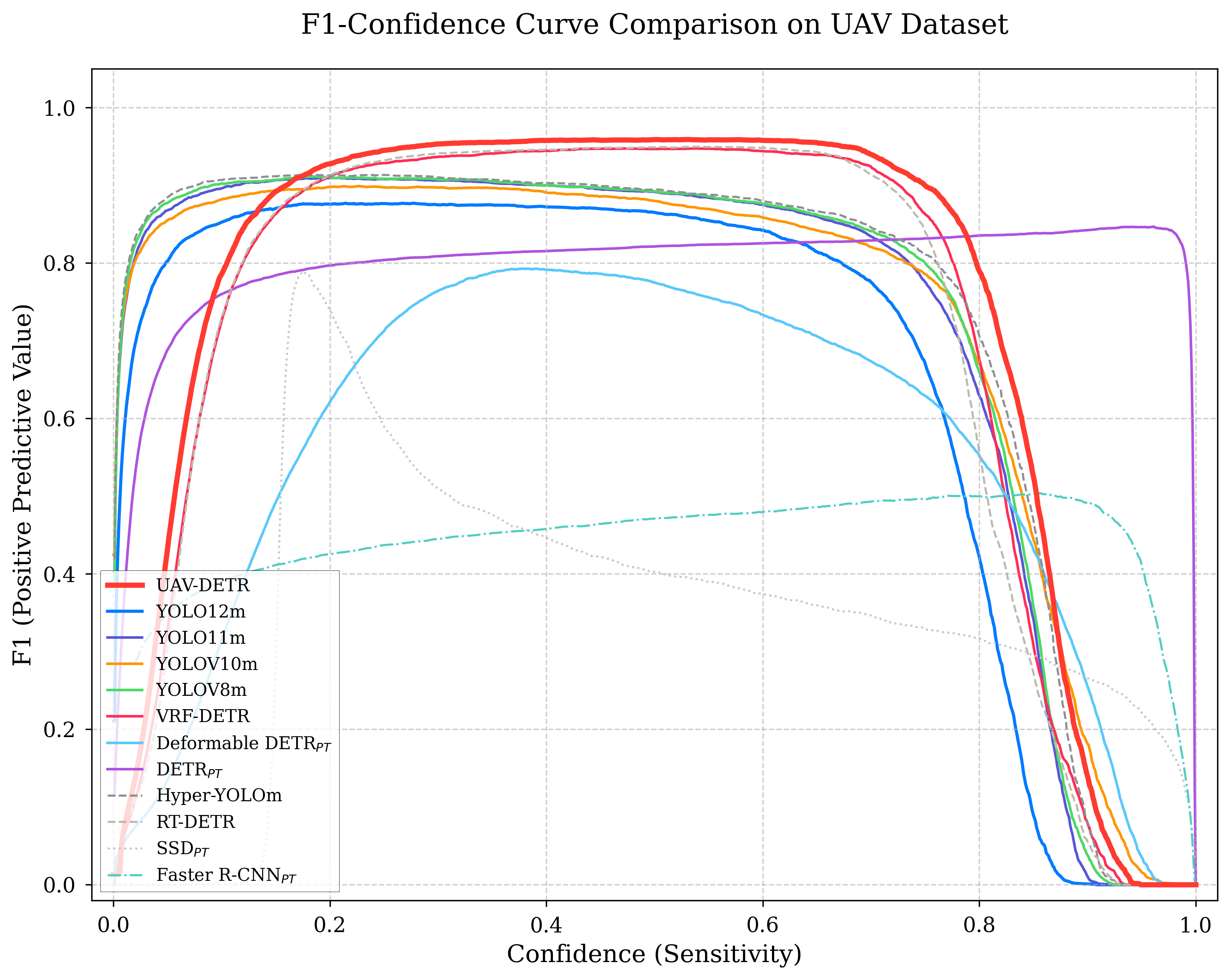}
		\centerline{(a) F1-Confidence Curve}
	\end{minipage}
	\hfill
	\begin{minipage}{0.49\linewidth}
		\centering
		\includegraphics[width=\linewidth]{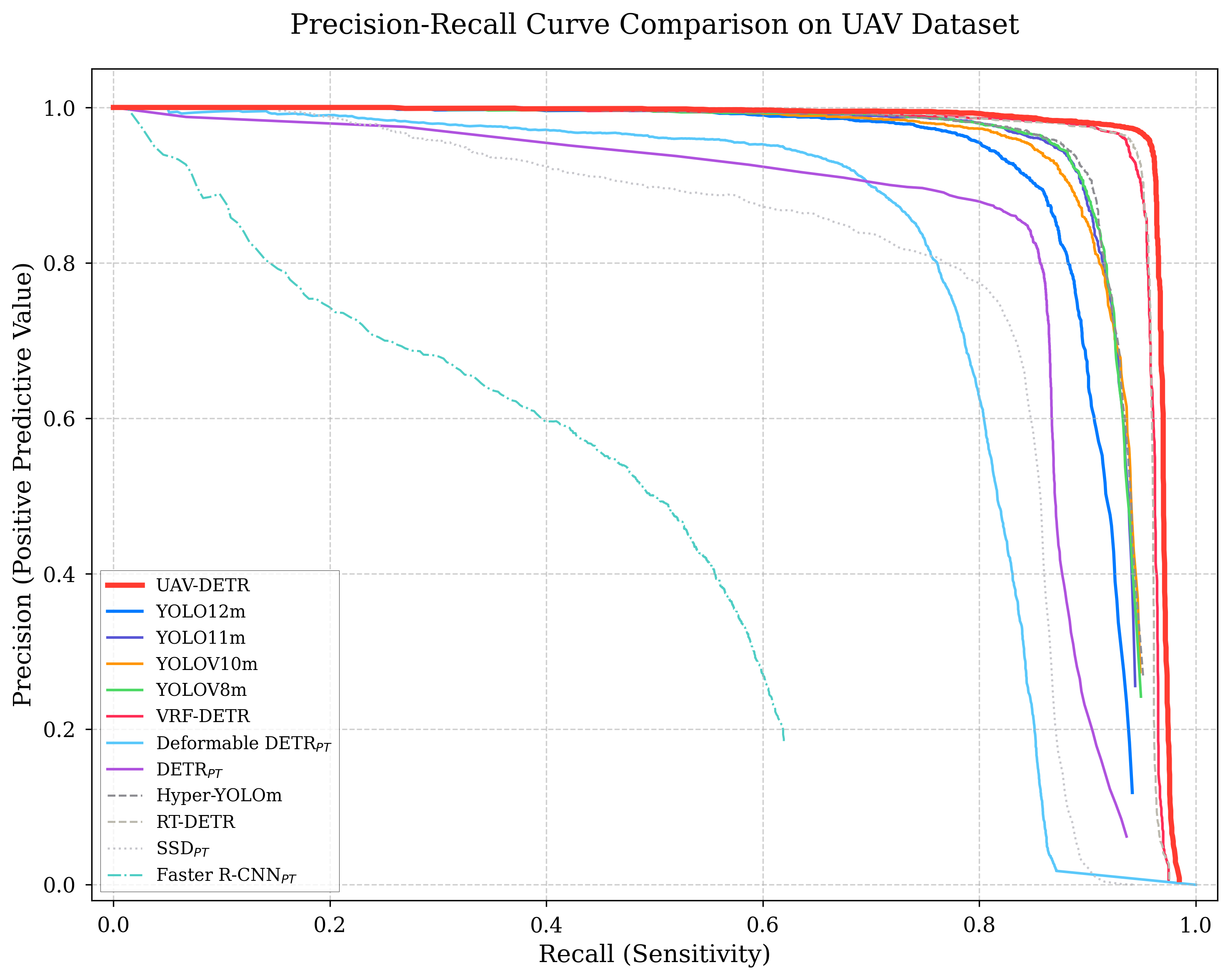}
		\centerline{(b) Precision-Recall Curve}
	\end{minipage}
	\caption{Qualitative comparison on the UAV dataset. The proposed UAV-DETR demonstrates superior stability and coverage in both (a) F1-Confidence and (b) Precision-Recall metrics.}
	\label{fig:curves}
\end{figure}

The visual comparison of the F1-Confidence and Precision-Recall curves in Fig.~\ref{fig:curves} offers an intuitive assessment of model robustness. As observed in the F1-Confidence metric, the proposed UAV-DETR (indicated by the bold red line) maintains a consistently high F1-score across a wide range of confidence thresholds, exhibiting a broader plateau compared to competitive models like RT-DETR and YOLO12m. It is worth noting that while models heavily reliant on pre-trained weights, such as DETR$_{\text{PT}}$ and Faster R-CNN$_{\text{PT}}$, appear to maintain higher F1-scores at extremely high confidence levels, this phenomenon is entirely attributable to the generic prior knowledge acquired from large-scale pre-training datasets. Crucially, in the broad and more practical intermediate confidence ranges, UAV-DETR significantly outperforms these pre-trained models by a large margin, demonstrating superior robustness.

Similarly, the Precision-Recall curve demonstrates that our method encompasses the largest area under the curve. While models denoted with the PT subscript depend on pre-training to achieve meaningful detection results—and often fail to effectively converge without it—UAV-DETR is trained entirely from scratch. Despite this strict and completely fair setting, our model not only avoids the convergence failure typical of standard Transformer detectors trained without prior weights, but also sustains superior precision in the high-recall region. In this region, early methods suffer from severe recall truncation and modern detectors experience a sharp precision decline. This characteristic indicates that our frequency-aware backbone and hybrid loss strategy effectively suppress false positives in complex backgrounds. Ultimately, this demonstrates that the performance gain of UAV-DETR stems from superior architectural design and domain-specific inductive bias rather than a reliance on massive generic data.

\begin{figure}[t]
	\centering
	\includegraphics[width=1.0\linewidth]{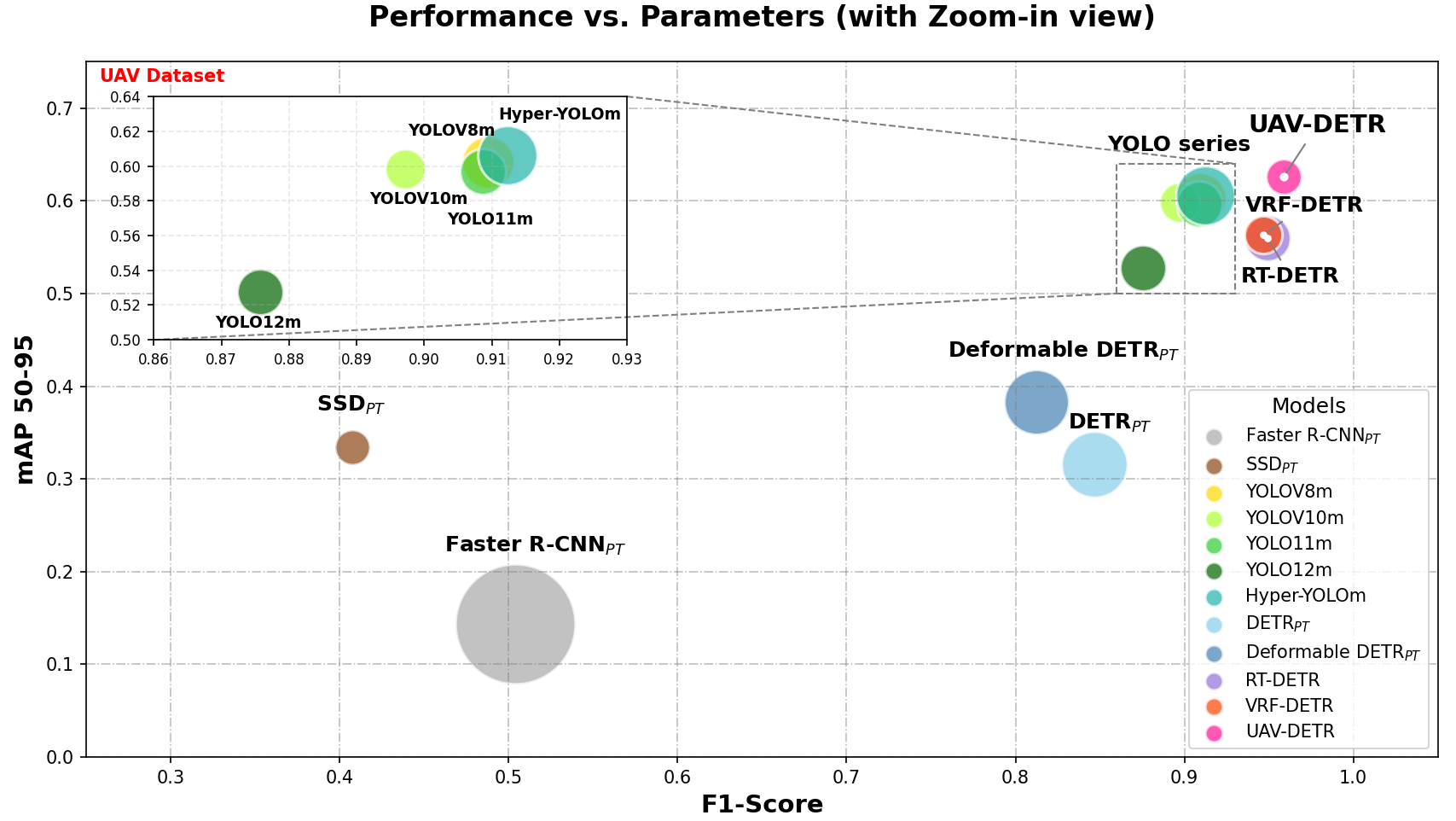}
	\caption{Comparison of Model Performance vs. Parameters. The x-axis and y-axis denote F1-Score and mAP$_{50:95}$, respectively, while the bubble size represents the number of parameters. Our UAV-DETR achieves the best trade-off, located in the top-right corner with a compact model size.}
	\label{fig:tradeoff}
\end{figure}

To further visualize the comprehensive trade-off between model complexity and detection capability, we map the performance landscape in Fig.~\ref{fig:tradeoff}. In this multidimensional scatter plot, the x-axis and y-axis denote F1-Score and mAP$_{50:95}$ respectively, while the bubble magnitude represents the parameter count (where smaller indicates lighter). Visually, UAV-DETR (highlighted in pink) distinctively occupies the optimal top-right position, signifying the highest simultaneous achievement in F1-Score and mAP. Crucially, the zoomed-in view reveals that despite maintaining a compact footprint ($\sim$11.9M parameters) comparable to the lightweight SSD$_{\text{PT}}$, our model delivers robust accuracy that rivals or even exceeds far larger architectures, validating the effectiveness of the proposed lightweight design.

\subsubsection{4.4.2 Quantitative Analysis}
\setlength{\parindent}{10pt}
The detailed numerical comparisons summarized in Table \ref{tab:uavdataset} demonstrate that UAV-DETR achieves a superior balance between detection precision and model complexity.

In terms of \textbf{detection accuracy}, our method outperforms all competing baselines across key metrics. Specifically, UAV-DETR attains the highest precision of 96.82\% and recall of 94.93\%, surpassing the second-best model, RT-DETR, by margins of 0.54\% and 1.30\%, respectively. More critically, under the rigorous mAP$_{50:95}$ metric, our model reaches 62.56\%, significantly outperforming advanced anchor-free models such as YOLO12m at 52.76\% and Hyper-YOLOm at 60.61\%. The substantial mAP$_{75}$ score of 71.08\% further confirms that the proposed ECFRFN architecture and the hybrid loss strategy, which integrates Inner-CIoU and NWD, significantly improve the geometric alignment and boundary regression for tiny objects. Crucially, this superior performance does not come at the cost of heavy computational overhead. 

Regarding the \textbf{efficiency metrics}, traditional models like Faster R-CNN$_{\text{PT}}$ and DETR$_{\text{PT}}$ suffer from excessive FLOPs and parameter counts exceeding 40M. While SSD$_{\text{PT}}$ achieves the lowest parameter count of 11.67M, its accuracy remains suboptimal with an mAP$_{50}$ of 78.16\%. In contrast, UAV-DETR establishes an optimal trade-off. With only 11.96M parameters, our model is approximately 53\% smaller than YOLOv8m and 40\% smaller than RT-DETR, yet it delivers significantly higher precision. Although VRF-DETR exhibits lower FLOPs, UAV-DETR outperforms it by 6.25\% in mAP$_{50:95}$ while maintaining a smaller overall footprint. This optimal trade-off is fundamentally driven by the WTConv-enhanced backbone, which drastically reduces parameter redundancy, coupled with the efficient feature processing of the ECFRFN and SWSA-IFI modules. Together, they successfully maximize the representational capacity for small targets within a highly compact budget.

\begin{table}[H]
	\centering
	\caption{Quantitative comparison of detection performance on the custom UAV dataset. The best results are highlighted in \textbf{bold}.}
	\label{tab:uavdataset}
	\vspace{4pt} % 【核心魔法1】：手动增加标题与表格之间的距离，防止标题“吃掉”线条
	\renewcommand{\arraystretch}{1.1} % 恢复到1.1，保证上下数据的呼吸感
	\setlength{\tabcolsep}{2pt} 
	\resizebox{\linewidth}{!}{ 
		\begin{tabular}{l @{\hspace{15pt}} cccccccc}
			\toprule[1.2pt] % 【核心魔法2】：强制顶线加粗到 1.2pt
			\textbf{Model} & $\boldsymbol{P}$(\%)$\uparrow$ & $\boldsymbol{R}$(\%)$\uparrow$ & $\boldsymbol{F1}$(\%)$\uparrow$ & \textbf{mAP}$_{\mathbf{50}}$(\%)$\uparrow$ & \textbf{mAP}$_{\mathbf{75}}$(\%)$\uparrow$ & \textbf{mAP}$_{\mathbf{50:95}}$(\%)$\uparrow$ & \textbf{FLOPs}(G)$\downarrow$ & \textbf{Params}$\downarrow$ \\
			\midrule[0.5pt] % 强制中间线保持 0.5pt 的细线状态
			Faster R-CNN$_{\text{PT}}$ & 53.65 & 47.51 & 50.40 & 43.62 & 6.04 & 14.41 & 401.7 & 136,689,024 \\
			SSD$_{\text{PT}}$ & 95.39 & 25.90 & 40.74 & 78.16 & 23.23 & 33.43 & 58.4 & \textbf{11,671,638} \\
			YOLOv8m      & 94.74 & 87.43 & 90.94 & 93.06 & 67.54 & 60.21 & 78.7 & 25,840,339 \\
			YOLOv10m     & 93.49 & 86.23 & 89.72 & 92.91 & 67.51 & 59.85 & 58.9 & 15,313,747 \\
			YOLO11m      & 94.51 & 87.50 & 90.87 & 92.97 & 67.02 & 59.69 & 67.6 & 20,030,803 \\
			YOLO12m      & 91.70 & 83.79 & 87.57 & 90.35 & 56.43 & 52.76 & 67.1 & 20,105,683 \\
			Hyper-YOLOm  & 94.55 & 88.13 & 91.23 & 93.41 & 68.56 & 60.61 & 103.1 & 33,336,307 \\
			DETR$_{\text{PT}}$ & 86.04 & 83.35 & 84.67 & 82.67 & 15.52 & 31.54 & 73.6 & 41,302,368 \\
			Def-DETR$_{\text{PT}}$ & 92.24 & 72.59 & 81.24 & 78.26 & 32.30 & 38.28 & 157.4 & 39,847,265 \\
			RT-DETR      & 96.28 & 93.63 & 94.94 & 95.45 & 58.55 & 55.95 & 56.9 & 19,873,044 \\
			VRF-DETR     & 96.10 & 93.27 & 94.66 & 95.46 & 61.19 & 56.31 & \textbf{44.2} & 13,537,896 \\
			\midrule[0.5pt] % 强制细线
			\textbf{UAV-DETR} & \textbf{96.82} & \textbf{94.93} & \textbf{95.87} & \textbf{96.58} & \textbf{71.08} & \textbf{62.56} & 66.7 & \textbf{11,962,040} \\
			\bottomrule[1.2pt] % 【核心魔法3】：强制底线也加粗到 1.2pt，与顶线呼应
		\end{tabular}
	}
\end{table}

\subsubsection{4.4.3 Visual Results}
\begin{figure}[H]
	\centering
	\includegraphics[width=\linewidth]{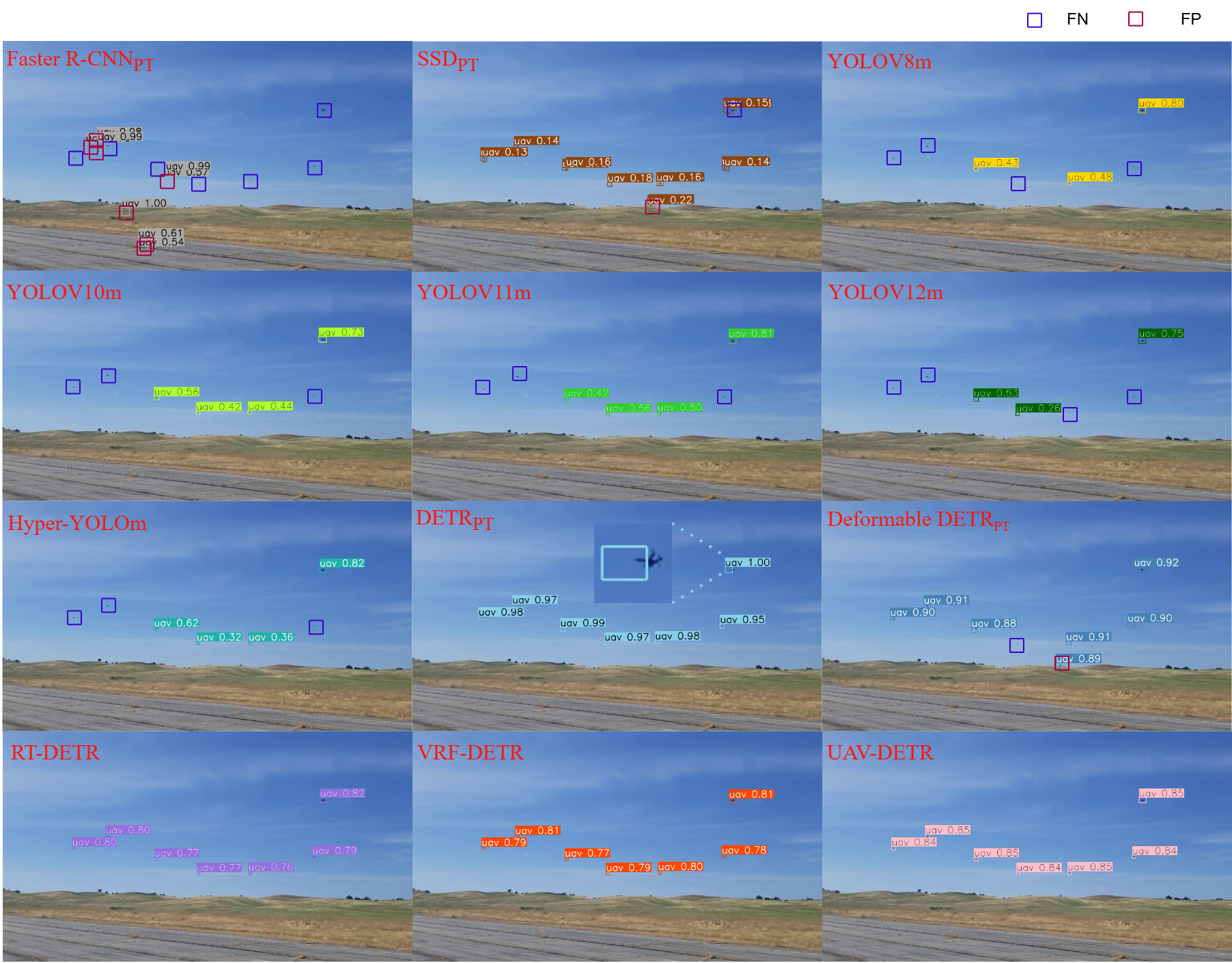}
	\caption{Visualization of detection results in an open field scenario. Purple and red boxes indicate missed detections and false alarms, respectively. The proposed UAV-DETR (bottom right) achieves precise localization of UAV targets with zero missed detections or false alarms..}
	\label{fig:visualization1}
\end{figure}

\begin{figure}[ht]
	\centering
	\includegraphics[width=\linewidth]{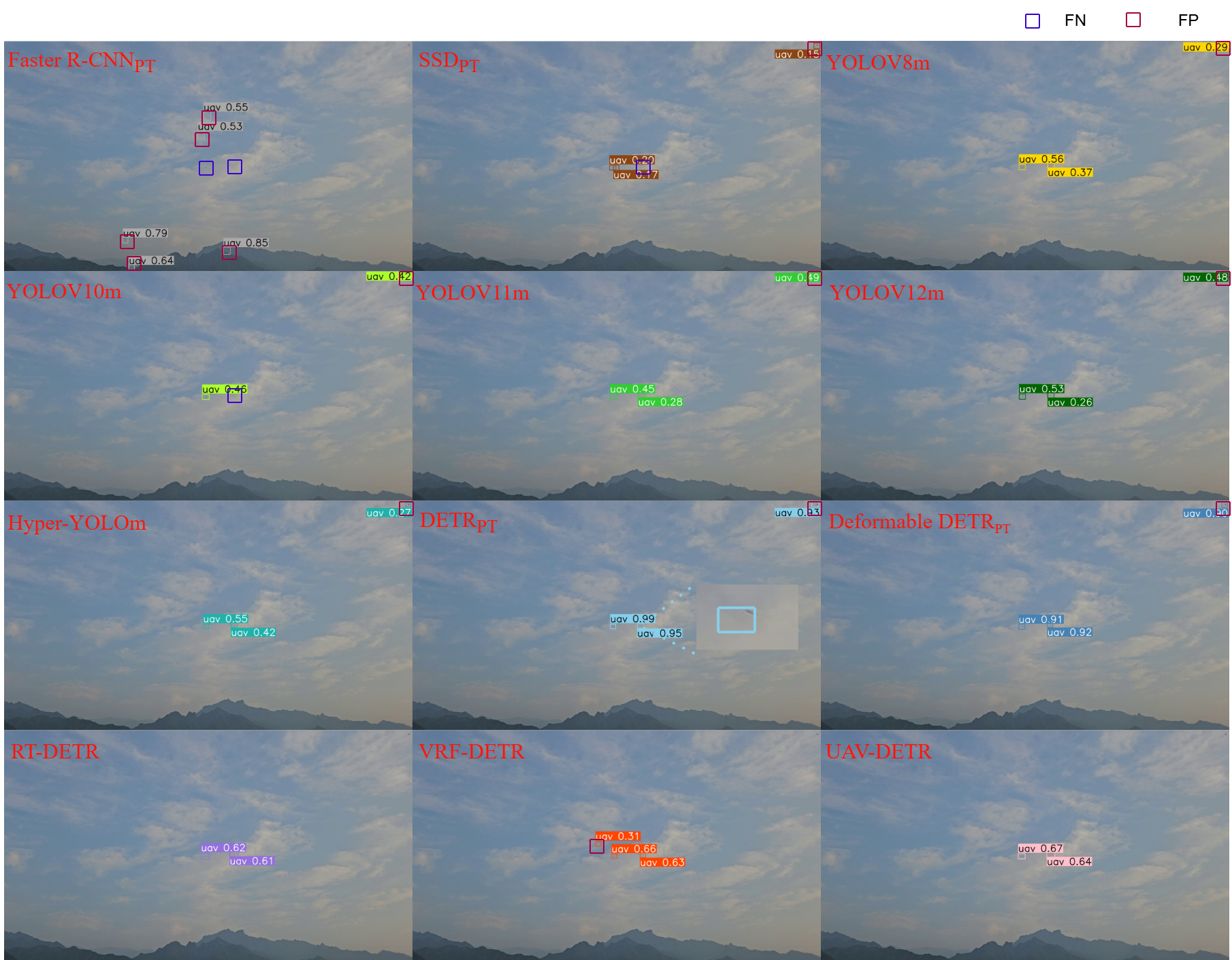}
	\caption{Visualization of detection results against a cloudy mountainous sky. The proposed UAV-DETR (bottom right) exclusively achieves flawless localization with elevated confidence scores, successfully avoiding both the bird-induced false alarms common in most baselines and other environmental clutter.}
	\label{fig:visualization2}
\end{figure}

\begin{figure}[ht]
	\centering
	\includegraphics[width=\linewidth]{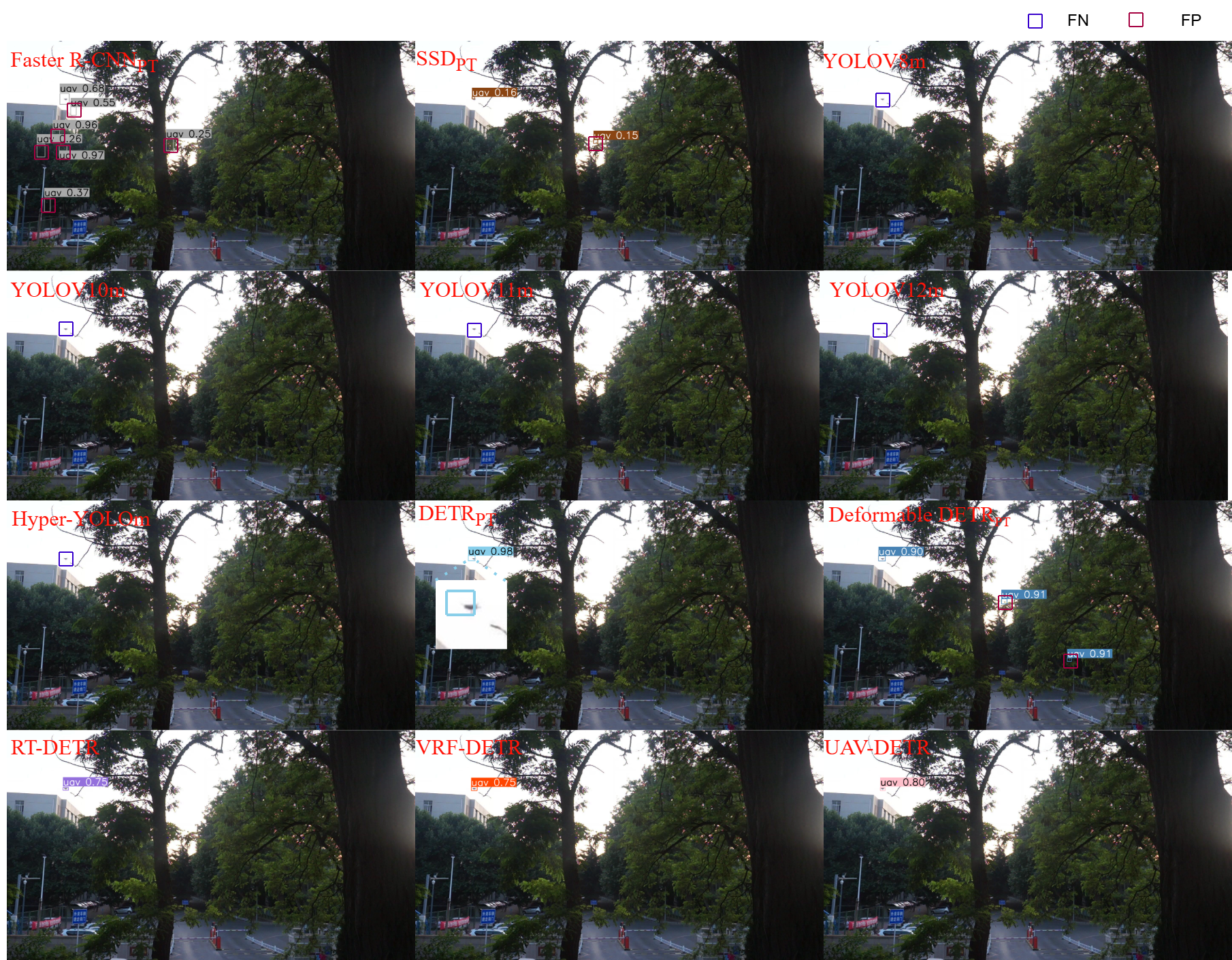}
	\caption{Visualization of detection results in a complex environment with heavy tree occlusion and structural clutter. The severe visual fragmentation causes a complete detection failure across all tested YOLO variants. In contrast, the proposed UAV-DETR (bottom right) successfully penetrates the foliage clutter, achieving accurate localization with the highest confidence score of 0.80.}
	\label{fig:visualization3}
\end{figure}

%这是三张可视化结果的图，每张图有12个子图。3*4，分别是ssd_pt、fastrcnn、yolov8、yolov10、yolov11、yolov12、hyperyolo、detr_pt、deformabledetr_pt、rtdetr、vrfdetr、uavdetr,fn表示漏检、fp表示误检，请在这里先介绍布局规则，然后再介绍一下结果，可以不用很详细，值得一提的是，detr的置信度很高，但是边界框是“歪的”，所以评估时指标不高，但是置信度高
Figures \ref{fig:visualization1}--\ref{fig:visualization3} visualize the qualitative detection results across diverse counter-UAV scenarios. Standard bounding boxes with confidence scores represent the models' raw predictions. To explicitly highlight detection failures, we overlay red boxes on false alarms (False Positives, FP) and use purple boxes to indicate missed targets (False Negatives, FN).

As depicted in Fig.~\ref{fig:visualization1}, the first scenario features multiple miniature drone targets in an open field, rigorously testing the models' capability to capture extreme small-scale features against natural backgrounds. Traditional CNNs and the advanced YOLO series, ranging from YOLOv8m to YOLO12m alongside Hyper-YOLOm, struggle noticeably with spatial resolution degradation and uniformly yield FNs for the smallest distant targets. Among the Transformer-based architectures utilizing pre-trained weights for baseline convergence, DETR$_{\text{PT}}$ yields severely misaligned bounding boxes despite high classification confidence scores near 1.0, which perfectly explains its suboptimal quantitative mAP. Deformable DETR$_{\text{PT}}$ slightly improves localization precision but still incurs one FN and one FP. In stark contrast, RT-DETR, VRF-DETR, and the proposed UAV-DETR successfully localize all targets without any FNs or FPs. Most notably, UAV-DETR attains the highest confidence scores among these flawless detectors, confirming its superior capability in precise geometric localization and overall detection robustness even against heavily pre-trained baselines.

Figure~\ref{fig:visualization2} illustrates a deceptive scenario under a cloudy mountainous sky, featuring two distant drones flying close to each other and a small bird situated far away in the upper right corner. Because the extreme distance reduces both the drones and the bird to visually similar dark spots, the bird acts as a severe biological distractor. Traditional CNNs struggle fundamentally: Faster R-CNN$_{\text{PT}}$ fails to detect any true targets and generates multiple false alarms, while SSD$_{\text{PT}}$ yields one missed detection, one false alarm, and redundant bounding boxes on a single true target. More critically, advanced architectures ranging from the entire YOLO series and Hyper-YOLOm to DETR$_{\text{PT}}$ and Deformable DETR$_{\text{PT}}$ fall into a common semantic trap by misclassifying the bird as a drone. Additionally, YOLOv10m incurs a missed detection, and DETR$_{\text{PT}}$ demonstrates its characteristic bounding box misalignment. Although VRF-DETR successfully avoids the bird distractor, it succumbs to complex atmospheric clutter, generating a false alarm elsewhere. Ultimately, only RT-DETR and the proposed UAV-DETR achieve flawless detection, accurately localizing both drones without any false alarms. Showcasing its clear superiority, UAV-DETR yields notably higher classification confidence, elevating the score of the left target from 0.62 in RT-DETR to 0.67. This confirms the exceptional capability of UAV-DETR to distinguish genuine mechanical targets from challenging biological and environmental distractors.

Furthermore, Fig.~\ref{fig:visualization3} presents a highly complex urban environment characterized by severe tree occlusion and heterogeneous background clutter. Detecting miniature drones through heavy foliage is a critical challenge, as the target is visually fragmented by branches and leaves. Under these extreme conditions, traditional detectors exhibit catastrophic degradation: Faster R-CNN$_{\text{PT}}$ successfully detects the target but suffers from numerous false alarms in the canopy, while SSD$_{\text{PT}}$ barely registers the drone with a critically low confidence alongside additional misclassifications. Strikingly, the entire suite of advanced YOLO architectures, spanning YOLOv8m to YOLO12m and Hyper-YOLOm, experiences a complete detection failure, uniformly missing the heavily obscured target. Among the pre-trained Transformer baselines, DETR$_{\text{PT}}$ detects the target but maintains its characteristic poor bounding box alignment, whereas Deformable DETR$_{\text{PT}}$ hallucinates multiple false alarms within the dense branches. Ultimately, only RT-DETR, VRF-DETR, and the proposed UAV-DETR successfully pierce through the visual fragmentation to accurately localize the drone. Demonstrating unparalleled robustness, UAV-DETR once again achieves the highest classification confidence, elevating the score to 0.80 compared to 0.75 in the baseline RT-DETR. This compelling visual evidence confirms that our proposed frequency-aware backbone and global attention mechanisms effectively filter out severe structural clutter to maintain precise and highly confident target focus.

Across all three visualized scenarios, a consistent architectural behavior emerges within the DETR family, particularly for the standard DETR$_{\text{PT}}$ and Deformable DETR$_{\text{PT}}$ methods. While these models routinely assign exceptionally high classification confidence scores frequently exceeding 0.90 to their predictions, their localized bounding boxes remain persistently skewed or loosely fitted relative to the tight physical boundaries of the miniature drones. Consequently, although these variants exhibit high semantic certainty regarding the presence and general vicinity of the targets, this inherent geometric misalignment severely penalizes their Intersection over Union (IoU) computation against the ground truth. This fundamental spatial discrepancy serves as the primary explanation for their suboptimal quantitative evaluation metrics, highlighting a critical disconnect between classification confidence and localization accuracy that our proposed modules successfully resolve.

\subsubsection{4.4.4 Visualization of Key Component Features}

\begin{figure}[H]
	\centering
	\includegraphics[width=\linewidth]{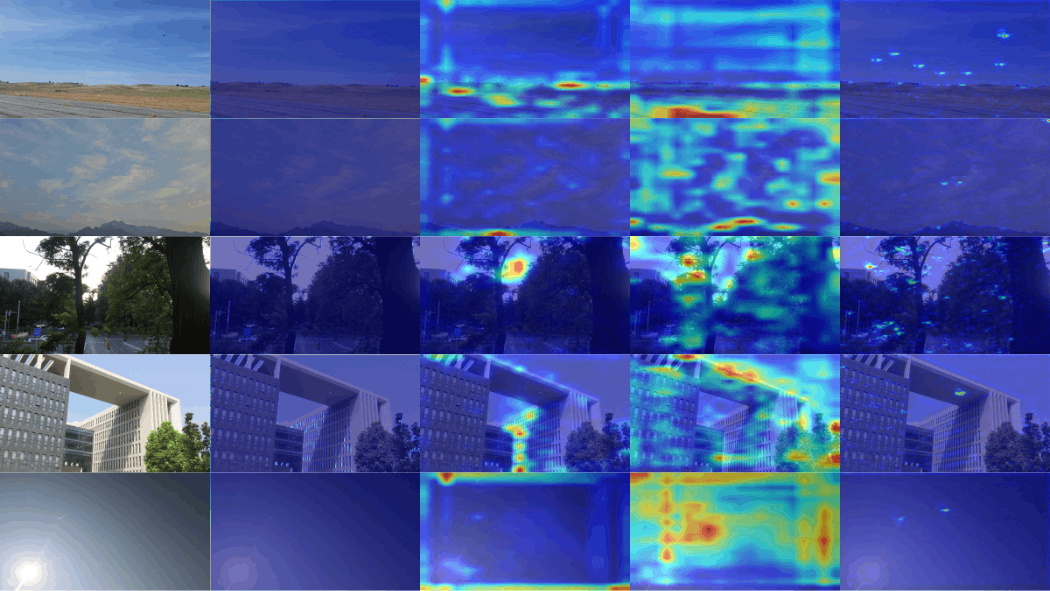} 
	\caption{Sequential feature heatmap visualization demonstrating the internal data flow of UAV-DETR. From left to right: (1) raw inputs, (2) low-level features, (3) WTConv-enhanced backbone outputs, (4) SWSA-IFI encoder outputs, and (5) final ECFRFN neck activations. The visual progression highlights the robust suppression of background clutter and the precise accentuation of miniature targets.}
	\label{fig:key_layers_heatmap}
\end{figure}

Visualizing the intermediate feature heatmaps extracted from sequential stages of the UAV-DETR pipeline intuitively demonstrates the internal mechanisms and the effectiveness of our specifically designed modules. Fig.~\ref{fig:key_layers_heatmap} presents these visualizations organized in a grid format to illustrate the progressive recalibration process across five distinct complex scenarios. The layout strictly follows the internal data flow: the five columns from left to right correspond to the raw input images, the initial low-level features from the primary ConvNormLayer, the high-frequency detail preservation after the WTConv-enhanced backbone, the global context focusing from the SWSA-IFI encoder, and the final noise-filtered activations from the ECFRFN neck at the $P_3$ resolution level.

The visual progression reveals a clear mechanism of noise suppression and target accentuation. The leftmost column displays miniature drones situated in highly challenging environments. Moving to the second column, the primary ConvNormLayer merely performs low-level pixel transformations with minimal semantic target awareness, resulting in uniformly weak activations. Subsequently, the WTConv-enhanced backbone, shown in the third column, successfully preserves high-frequency structural details and prevents the severe information loss typical of standard downsampling. However, substantial background clutter, such as tree foliage and architectural edges, is also prominently activated alongside the targets. The SWSA-IFI encoder addresses this interference by establishing global context awareness and actively shifting the network attention towards salient regions, as seen in the fourth column. While this mechanism significantly enhances the semantic focus on potential targets, it occasionally highlights prominent background distractors due to its broad receptive field. Finally, the fifth column showcases the culmination of the feature refinement process. Through selective boundary aggregation and cross-scale fusion within the ECFRFN module, the remaining structural background noise is completely filtered out. The final heatmaps present highly localized and intense activations concentrated exclusively on the true drone targets, appearing as distinct sharp spots. This sequential visualization confirms that each proposed component contributes indispensably to isolating tiny aerial targets from severe environmental interference.

% 插入关键层特征可视化图

%-------------------------开源数据集--------------------------------------
\subsubsection{4.4.5. Generalization Verification on Public Benchmark}
\begin{figure}[H]
	\centering
	\includegraphics[width=1.0\linewidth]{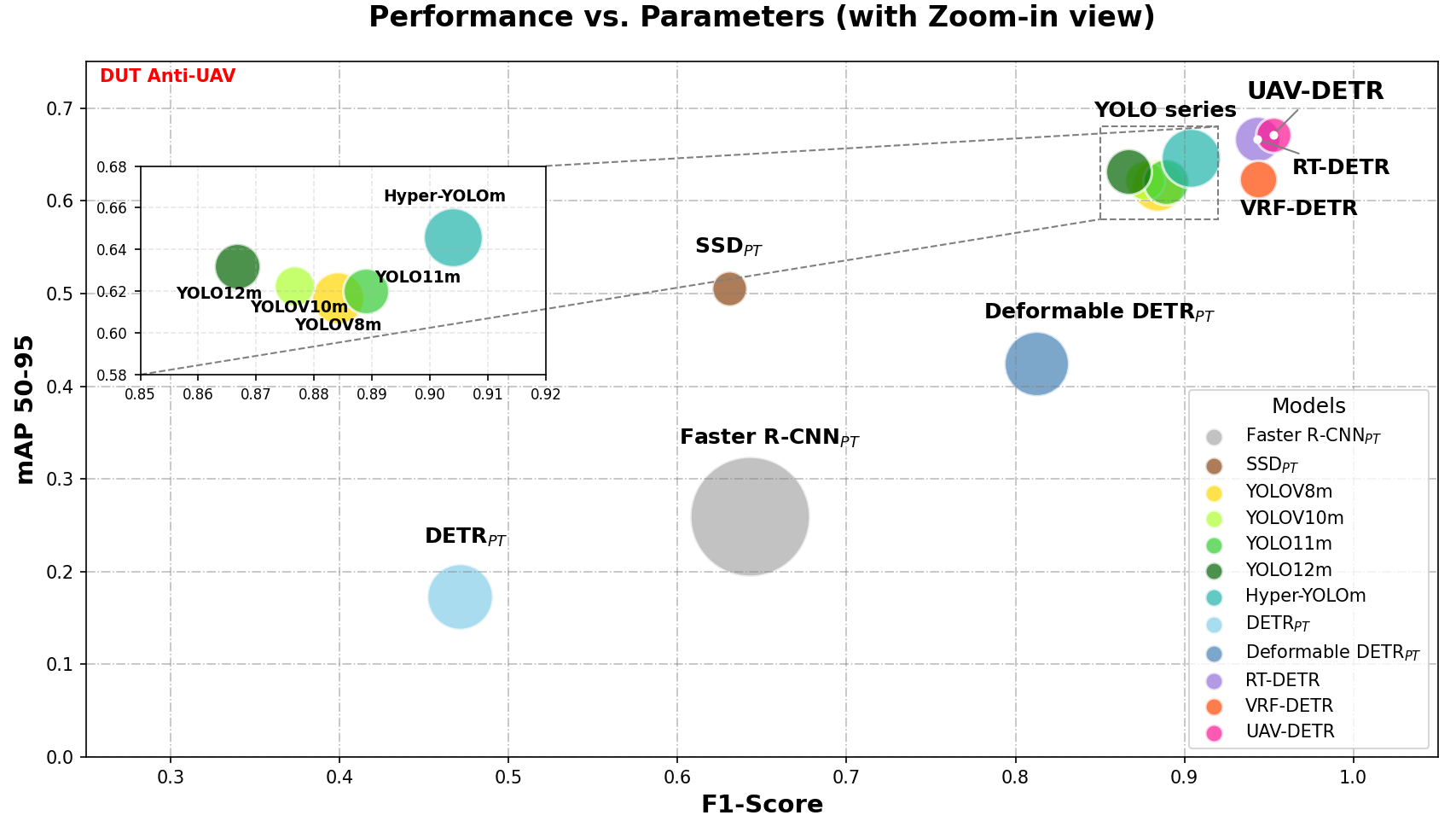}
	\caption{Performance comparison on the DUT Anti-UAV dataset. UAV-DETR maintains its leading position, demonstrating strong generalization capabilities.}
	\label{fig:dut_perf}
\end{figure}

\setlength{\parindent}{10pt}
To promote reproducibility and facilitate further research within the community, we have fully open-sourced our source code and the constructed dataset, which are accessible via the GitHub link provided in the abstract. To further rigorously validate the generalization capability of the proposed method across different data distributions beyond our self-collected samples, we extended our evaluation to the publicly available DUT Anti-UAV dataset. Fig.~\ref{fig:dut_perf} provides an intuitive visualization of this robust generalization. As visualized, even on this external benchmark with varying environmental characteristics, UAV-DETR maintains a distinct performance advantage over competing methods, validating that our model has not overfitted to the self-constructed dataset.

The detailed numerical results, listed in Table \ref{tab:dut}, further corroborate this superiority. Consistently, UAV-DETR achieves state-of-the-art performance across all reported metrics. Notably, it attains an mAP$_{50:95}$ of 67.15\%, surpassing the highly competitive RT-DETR at 66.67\% and the advanced YOLO12m at 63.19\%. The significant lead in Precision, reaching 97.09\%, further indicates that our frequency-aware design effectively distinguishes UAV targets from complex backgrounds in diverse environments, minimizing false alarms.

Regarding the balance between efficiency and accuracy, UAV-DETR demonstrates exceptional adaptability for practical deployment. As shown in the efficiency metrics, our model maintains an extremely lightweight footprint with only 11.8 M parameters. This is comparable to the 11.6 M parameters of the lightweight SSD$_{\text{PT}}$, yet it offers a massive improvement in accuracy, with mAP$_{50:95}$ increased by approximately 16.6\%. Compared to VRF-DETR, which focuses purely on low FLOPs, our method provides a superior trade-off, delivering a 4.86\% higher mAP$_{50:95}$ with a smaller model size, thereby proving its suitability for resource-constrained platforms.

\begin{table}[H]
	\centering
	\caption{Quantitative comparison of detection performance on the public DUT-ANTI-UAV benchmark. The best results are highlighted in \textbf{bold}.}
	\label{tab:dut}
	\vspace{4pt} 
	\renewcommand{\arraystretch}{1.1} % 恢复到1.1，保证上下数据的呼吸感
	\setlength{\tabcolsep}{2pt} 
	\resizebox{\linewidth}{!}{ 
		\begin{tabular}{l @{\hspace{15pt}} cccccccc}
			\toprule[1.2pt] % 【核心魔法2】：强制顶线加粗到 1.2pt
			\textbf{Model} & $\boldsymbol{P}$(\%)$\uparrow$ & $\boldsymbol{R}$(\%)$\uparrow$ & $\boldsymbol{F1}$(\%)$\uparrow$ & \textbf{mAP}$_{\mathbf{50}}$(\%)$\uparrow$ & \textbf{mAP}$_{\mathbf{75}}$(\%)$\uparrow$ & \textbf{mAP}$_{\mathbf{50:95}}$(\%)$\uparrow$ & \textbf{FLOPs}(G)$\downarrow$ & \textbf{Params}$\downarrow$ \\
			\midrule
			Faster R-CNN$_{\text{PT}}$ & 77.62 & 54.83 & 64.27 & 49.89 & 25.82 & 26.01 & 401.7 & 136,689,024 \\
			SSD$_{\text{PT}}$ & 95.74 & 47.04 & 63.08 & 80.12 & 54.64 & 50.50 & 58.4 & \textbf{11,671,638} \\
			YOLOv8m      & 92.84 & 84.38 & 88.41 & 90.61 & 68.59 & 61.66 & 78.7 & 25,840,339 \\
			YOLOv10m     & 92.63 & 83.21 & 87.66 & 91.14 & 69.62 & 62.25 & 58.9 & 15,313,747 \\
			YOLO11m      & 94.21 & 84.14 & 88.89 & 90.99 & 70.37 & 62.01 & 67.6 & 20,030,803 \\
			YOLO12m      & 93.03 & 81.11 & 86.67 & 90.48 & 71.43 & 63.19 & 67.1 & 20,105,683 \\
			Hyper-YOLOm   & 94.12 & 86.95 & 90.39 & 92.89 & 72.43 & 64.59 & 103.1 & 33,336,307 \\
			DETR$_{\text{PT}}$ & 62.06 & 37.96 & 47.11 & 41.94 & 12.60 & 17.33 & 73.6 & 41,302,368 \\
			Def-DETR$_{\text{PT}}$ & 92.24 & 72.59 & 81.24 & 78.36 & 41.04 & 42.48 & 157.4 & 39,847,265 \\
			RT-DETR      & 95.69 & 92.92 & 94.29 & 95.59 & 75.50 & 66.67 & 56.9 & 19,873,044 \\
			VRF-DETR     & 96.86 & 92.00 & 94.37 & 95.43 & 69.75 & 62.29 & \textbf{44.2} & 13,537,896 \\
			\midrule[0.5pt]
			\textbf{UAV-DETR} & \textbf{97.09} & \textbf{93.50} & \textbf{95.26} & \textbf{96.17} & \textbf{75.71} & \textbf{67.15} & 65.2 & \textbf{11,886,780} \\
			\bottomrule[1.2pt] % 【核心魔法3】：强制底线也加粗到 1.2pt，与顶线呼应
		\end{tabular}
	}
\end{table}
%-------------------------消融实验--------------------------------------
\subsection{4.5 Ablation Study}
\setlength{\parindent}{10pt}
%这是在uavdataset上面做的消融实验，其中O为baseline RTDETR O+A为替换损失函数为inner_ciou_nwd.O+A+B为O+A基础上加入ecrn，以此类推。O+A+B+C为引入WTConv，O+A+B+C+D为最终的加上SWSA-IFI。做O+A+B+D是因为O+A+B+C指标不太好，事实证明是C+D一起起的作用更大，所以C有必要加。
To systematically verify the efficacy of each proposed component and dissect their individual contributions to the overall performance, we conducted a comprehensive ablation study on the custom UAV dataset. We adopt the standard RT-DETR as the baseline (denoted as Model O) and progressively integrate the following modules:
\begin{itemize}
	\item \textbf{A}: The Hybrid Loss strategy (Inner-CIoU + NWD).
	\item \textbf{B}: The Efficient Cross-Scale Feature Recalibration and Fusion Network (ECFRFN).
	\item \textbf{C}: The Frequency-aware Backbone (WTConv Blocks).
	\item \textbf{D}: The Sliding Window Self-Attention Encoder (SWSA-IFI).
\end{itemize}

\begin{figure}[ht]
	\centering
	\includegraphics[width=\linewidth]{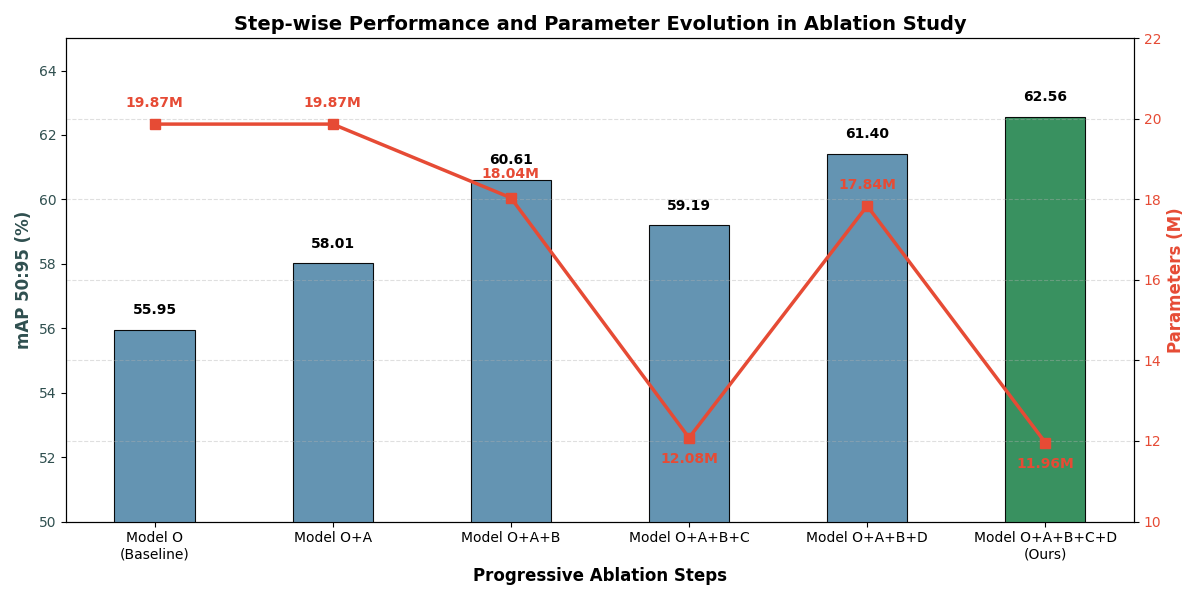}
	% 【修复处】：图注中的称呼也统一为 Model O+A+B+C+D
	\caption{Progressive ablation study evaluating the trade-off between detection accuracy and model complexity. The bar chart (left y-axis) denotes the \textbf{mAP}$_{\mathbf{50:95}}$ metric ($\uparrow$), while the line graph (right y-axis) tracks the model parameter count ($\downarrow$). The incremental steps demonstrate the impact of each component, with the final configuration (Model O+A+B+C+D) representing our proposed UAV-DETR, achieving the optimal balance between high precision and lightweight design.}
	\label{fig:ablation}
\end{figure}

We first visualize the evolution of model performance versus complexity in Fig.~\ref{fig:ablation}. The dual-axis chart reveals a clear optimization trajectory: while the detection accuracy (represented by the bar chart) exhibits a steady upward trend across the initial steps, a pivotal shift occurs upon the introduction of the WTConv backbone (Model O+A+B+C). Here, the parameter count (represented by the line chart) demonstrates a sharp decline, signifying a massive reduction in model redundancy. Crucially, the final configuration (Model O+A+B+C+D) achieves a distinct peak in accuracy while simultaneously occupying the lowest valley in parameter magnitude, visually confirming the effectiveness of our lightweight yet high-performance design strategy.

The detailed numerical comparisons are enumerated in Table \ref{tab:ablation}. Initially, replacing the original loss function with our Hybrid Loss strategy (Model O+A) yields immediate improvements. As evidenced by the transition from Row 1 to Row 2, the introduction of Inner-CIoU and NWD boosts mAP$_{50:95}$ from 55.95\% to 58.01\%. Notably, mAP$_{75}$ sees a significant jump of 4.16\%, validating that the hybrid loss effectively enhances geometric alignment accuracy for small targets. Subsequently, incorporating the ECFRFN module (Model O+A+B) to replace the standard neck further elevates mAP$_{50:95}$ to 60.61\%, demonstrating that the SBA mechanism successfully filters out background noise during multi-scale feature interaction.

To investigate the optimal balance between lightweight design and semantic representation, we further analyzed the specific roles of module C, the frequency-aware backbone, and module D, the global attention mechanism. Restructuring the backbone with our proposed WTConv Blocks (Model O+A+B+C) successfully achieves the primary goal of lightweight design, drastically reducing the parameter count from 18.04 M to 12.08 M. However, as observed in Row 4, this aggressive compression leads to a slight degradation in mAP$_{50:95}$, which drops to 59.19\%, suggesting that significantly reducing channel redundancy may entail a minor loss of semantic information. Conversely, adding only the attention module (Model O+A+B+D) improves accuracy but retains a higher computational burden of 17.84 M parameters. Most critically, when both modules are integrated (Model O+A+B+C+D), a remarkable synergy is achieved. As shown in the final row, the model attains the highest performance with an mAP$_{50:95}$ of 62.56\% while maintaining the lowest parameter count of 11.96 M. This confirms that the robust contextual features captured by the SWSA-IFI encoder effectively compensate for the semantic capacity reduced by the lightweight WTConv Blocks, resulting in an optimal trade-off between efficiency and precision.

% （表格代码保持上一版的最终形态即可，表头有 Model，行名用 O、O+A 等是学术界通用的清爽写法）

\begin{table}[H]
	\centering
	\caption{ Ablation Study. The best results are highlighted in \textbf{bold}.}
	\vspace{4pt}
	\label{tab:ablation}
	\renewcommand{\arraystretch}{1} 
	\setlength{\tabcolsep}{2pt} 
	\resizebox{\linewidth}{!}{ 
		\begin{tabular}{l @{\hspace{15pt}} cccccccc}
			\toprule[1.2pt]
			\textbf{Model} & $\boldsymbol{P}$(\%)$\uparrow$ & $\boldsymbol{R}$(\%)$\uparrow$ & $\boldsymbol{F1}$(\%)$\uparrow$ & \textbf{mAP}$_{\mathbf{50}}$(\%)$\uparrow$ & \textbf{mAP}$_{\mathbf{75}}$(\%)$\uparrow$ & \textbf{mAP}$_{\mathbf{50:95}}$(\%)$\uparrow$ & \textbf{FLOPs}(G)$\downarrow$ & \textbf{Params}$\downarrow$ \\
			\midrule
			O & 96.28 & 93.63 & 94.94 & 95.45 & 58.55 & 55.95 & \textbf{56.9} & 19,873,044 \\
			O+A & 95.95 & 94.18 & 95.05 & 95.76 & 62.71 & 58.01 & \textbf{56.9} & 19,873,044 \\
			O+A+B & 96.51 & 94.60 & 95.54 & 96.06 & 67.61 & 60.61 & 79.7 & 18,038,324 \\
			O+A+B+C & 96.22 & 94.44 & 95.32 & 95.54 & 65.03 & 59.19 & 65.2 & 12,082,484 \\
			O+A+B+D & 96.81 & 94.89 & 95.84 & 96.23 & 68.74 & 61.40 & 79.7 & 17,842,620 \\
			O+A+B+C+D & \textbf{96.82} & \textbf{94.93} & \textbf{95.87} & \textbf{96.58} & \textbf{71.08} & \textbf{62.56} & 66.7 & \textbf{11,962,040} \\
			\bottomrule[1.2pt]
		\end{tabular}
	}
\end{table}
\subsubsection{4.6 Discussion on Algorithm Failures and Limitations}
\setlength{\parindent}{10pt}
% 本章讨论算法失效的情况以及局限性，失效从主要从误检和漏检两种情况来说，图1是uavdetr误把鸟当作无人机的情况，虽然算法有时可以成功排除鸟的干扰，但是当无人机目标与鸟类成像高度相似，且相距不远时，依然会时而被干扰。图2是漏检的情况，主要是无人机目标与复杂的建筑物背景几乎融在一起，难以检测成功。局限性部分，我们的算法虽然参数量较小、性能优异，但是由于进行了特征重校准,以及其他操作，计算量相比baseline增加了9.8Fflos(17.2%)，需要依靠额外的剪枝和量化等步骤来减少。
Analyzing the failure cases and inherent limitations of UAV-DETR provides crucial insights for future optimizations, despite its state-of-the-art performance in various scenarios. Our empirical investigations reveal two predominant failure modes: FP induced by morphological distractors and FN caused by severe environmental camouflage.

\begin{figure}[H]
	\centering
	\begin{minipage}{0.48\linewidth}
		\centering
		\includegraphics[width=\linewidth]{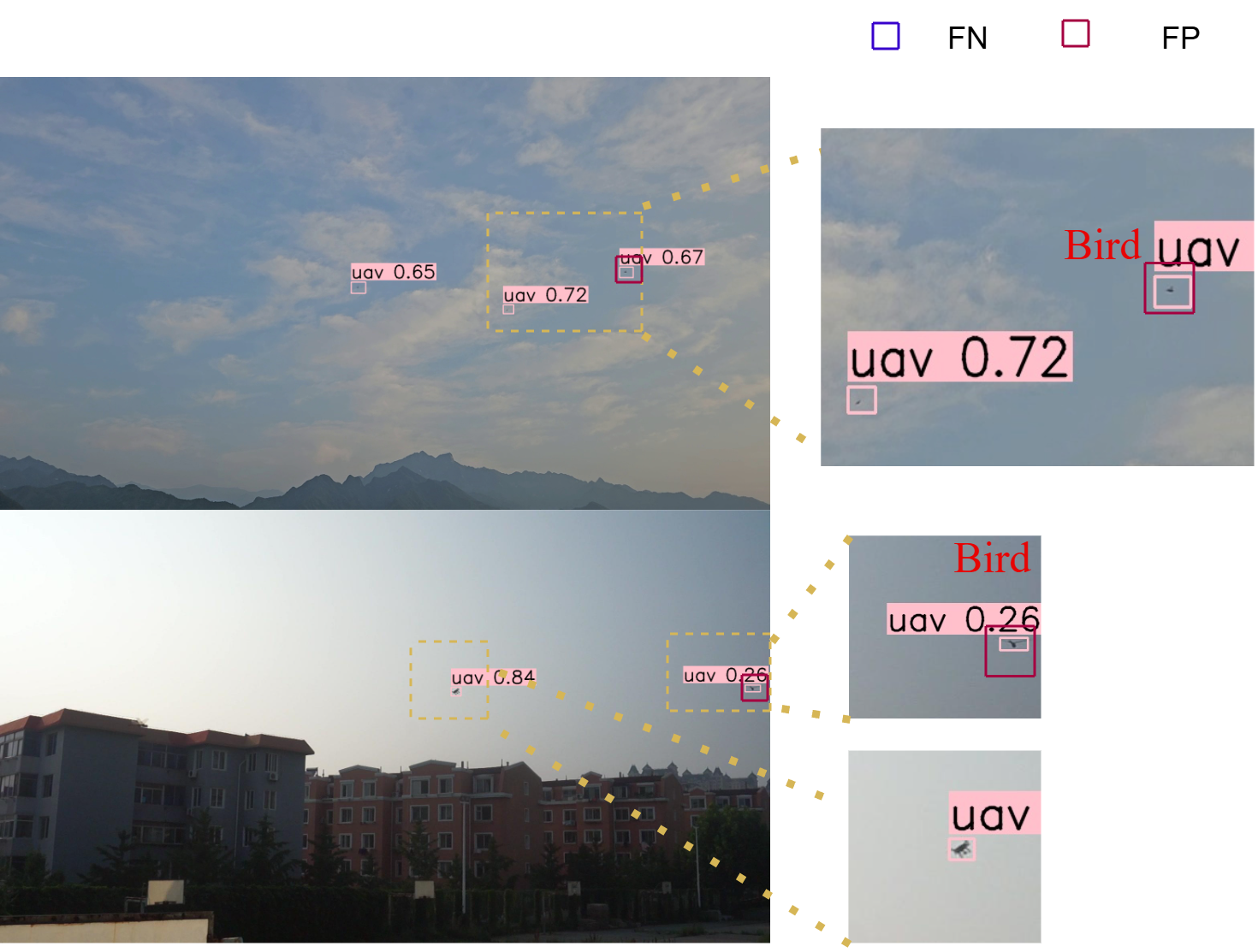}
		\centerline{(a) False Positives (FP)}
	\end{minipage}
	\hfill
	\begin{minipage}{0.48\linewidth}
		\centering
		\includegraphics[width=\linewidth]{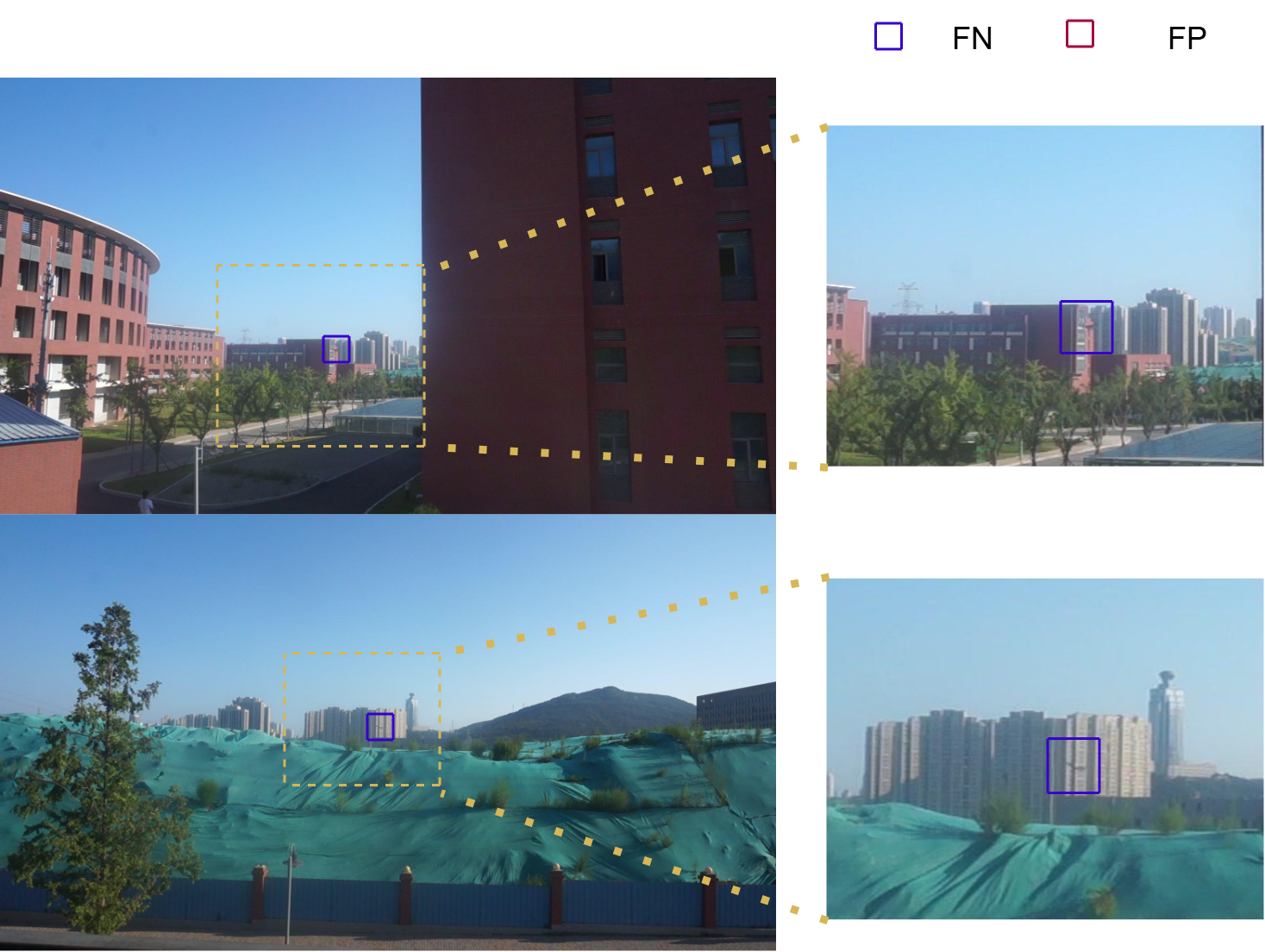}
		\centerline{(b) False Negatives (FN)}
	\end{minipage}
	\caption{Representative failure cases visualized by UAV-DETR on the custom UAV dataset. (a) FP where morphological similarities with birds lead to misclassification. (b) FN caused by miniature targets visually blending into complex urban architectural backgrounds.}
	\label{fig:unvisualization}
\end{figure}

Figure~\ref{fig:unvisualization}(a) systematically illustrates typical FP instances where flying birds are misclassified as drones. Although UAV-DETR excels at distinguishing biological distractors in standard environments, occasional misclassifications still persist when a bird exhibits extreme morphological similarity to a drone and is situated in close spatial proximity to actual targets. This specific perceptual ambiguity remains a challenging bottleneck under low-resolution conditions. Conversely, Fig.~\ref{fig:unvisualization}(b) highlights examples of missed detections primarily resulting from severe visual blending. In these urban environments, the miniature drone often visually merges with the highly complex and textured architectural background. The corresponding lack of distinct contrast hinders the model's ability to extract discriminative object boundaries, ultimately leading to FN.

Beyond environmental vulnerabilities, a key architectural limitation resides in the framework's increased computational overhead. Although UAV-DETR maintains a compact parameter count and achieves superior accuracy, the feature recalibration and fusion mechanisms introduce inevitable computational demands. Specifically, the overall computational volume increases by 9.8 GFLOPs, representing a 17.2\% overhead compared to the baseline RT-DETR. Consequently, future work must focus on subsequent optimization strategies, such as network pruning and weight quantization, to satisfy strict hardware constraints on ultra-low-power edge devices.

%--------------------------结论------------------------------------------------
\section{5. Conclusion}
\setlength{\parindent}{10pt}

In this paper, we proposed UAV-DETR, an efficient and robust object detection framework tailored to address the critical challenges of extreme scale variations, miniature target sizes, and complex background interference inherent in counter-UAV scenarios. By synergistically integrating a WTConv-enhanced backbone, an SWSA-IFI encoder, an ECFRFN neck architecture, and a hybrid InnerCIoU-NWD loss function, the proposed method significantly enhances multi-scale feature representation and geometric alignment for small aerial targets.

Comprehensive evaluations on a custom UAV dataset and the public DUT-ANTI-UAV benchmark validate the effectiveness and generalization capabilities of the proposed framework. UAV-DETR consistently outperforms 11 state-of-the-art detectors, including the recent YOLOV8m–YOLO12m series and advanced DETR variants. Specifically, it achieves an F1-Score of 95.87\% and an mAP$_{50:95}$ of 62.56\% on the custom dataset. This strong performance translates well to the DUT-ANTI-UAV dataset, yielding an F1-Score of 95.26\% and an mAP$_{50:95}$ of 67.15\%, demonstrating its robustness in mitigating false detections amid severe background clutter. Furthermore, detailed ablation studies confirm the individual contributions and synergistic effects of the proposed modules. The progressive integration of these components systematically improves detection accuracy, elevating the baseline mAP$_{50:95}$ from 55.95\% to 62.56\%. Concurrently, the network complexity is significantly reduced. UAV-DETR maintains a highly compact parameter footprint of 11.96 M—an approximate 40\% reduction from the 19.87 M baseline—thereby establishing an optimal trade-off between detection precision and model lightweighting.

In the future, while this study primarily validates UAV-DETR within counter-UAV scenarios, our failure analysis reveals that extreme morphological distractors and severe environmental camouflage still pose perceptual challenges. Additionally, the advanced feature recalibration inevitably introduces a 17.2\% increase in computational overhead. Consequently, our ongoing research will unfold in two primary directions. First, to neutralize this computational burden and satisfy the stringent low-latency constraints of real-world defense systems, we will explore hardware-aware optimization strategies, such as network pruning and weight quantization, to facilitate seamless deployment on ultra-low-power edge computing platforms (e.g., RK3588). Second, building upon this robust foundation, we aim to integrate advanced object tracking algorithms to handle the highly dynamic nature of multi-target drone swarms, marking a crucial step toward developing an autonomous, intelligent system capable of real-time combat intent recognition \cite{INTENTION}.
%----------------------------参考文献------------------------------------------
%\begin{thebibliography}{1}
%\bibitem{i}
%Cooler A.S.: Binary Flow Systems. \textit{J. Fluid Mech} \textbf{999}:999--996, 1999.
%
%\bibitem{ii}
%Icer D.F., Adams J.A.: Mathematical Elements for Computer Simulation.
%McGraw-Hill, NY 1977.
%
%\bibitem{iii}
%Nygus G.: Numerical Analysis Using Finite Element Method.
%\textit{PhD Thesis}, NTU Mech. Eng. Dept., Lagos, 1983.
%
%\bibitem{iv}
%ASTRA 2025, \textsf{\scriptsize{www.iist.ac.in/astra} }
%
%\end{thebibliography}

\bibliographystyle{unsrt}  % 参考文献样式
\bibliography{reference} 
\end{document}